\documentclass[12pt,a4paper,reqno]{amsart}
\usepackage[pdftex]{hyperref}

\usepackage[tight,footnotesize]{subfigure}
\usepackage[pdftex]{graphicx}

\usepackage{threeparttable}
\usepackage{multirow}

\usepackage{amssymb}
\usepackage{amscd}
\usepackage{enumerate}
\usepackage{graphicx}
\usepackage{siunitx}
\usepackage{tikz-cd}
\usepackage{bm}

\usepackage{bbm}

\usepackage{url}

\usepackage[ruled, commentsnumbered]{algorithm2e}

\numberwithin{equation}{section}

\usepackage{mathtools}%                  http://www.ctan.org/pkg/mathtools
\usepackage[tableposition=top]{caption}% http://www.ctan.org/pkg/caption
\usepackage{booktabs,dcolumn}%           http://www.ctan.org/pkg/dcolumn + http://www.ctan.org/pkg/booktabs

\DeclareFontFamily{OT1}{rsfs}{}
\DeclareFontShape{OT1}{rsfs}{n}{it}{<-> rsfs10}{}
\DeclareMathAlphabet{\mathscr}{OT1}{rsfs}{n}{it}

\theoremstyle{plain}

\theoremstyle{definition}

\usepackage[absolute,overlay]{textpos}

\begin{document}

\title[Nearest Constrained Subspace Classifier]{Intrinsic Dimension Estimation via Nearest Constrained Subspace Classifier}

\author{Liang Liao, Yanning Zhang, Stephen John Maybank, Zhoufeng Liu
%Liang Liao$^{1}$ and Stephen John Maybank$^{2}$
%\vspace{0.5em}\\
%$^{1}~$Zhongyuan University of Technology, ~~$^{2}~$Birkbeck, University of London
%\vspace{0.5em}\\
%{\MakeUppercase{liaoliangis@126.com}},\textit{~~~~} 
%{\MakeUppercase{sjmaybank@dcs.bbk.ac.uk}} 
}
%\address{China}
%\email{liaoliangis@126.com}

%\email{}

%\subjclass[2010]{37P99}

\begin{textblock}{12}(3.2, 4.2)
\MakeUppercase{\footnotesize liaoliangis@126.com (L. Liao)}, \MakeUppercase{\footnotesize sjmaybank@dcs.bbk.ac.uk (S. J. Maybank)}

\end{textblock}

\begin{abstract}  
\vspace{2em}
We consider the problems of classification and intrinsic dimension estimation on image data.
A new subspace based classifier is proposed for supervised classification or intrinsic dimension estimation. The distribution of the data in each class is modeled by a union of
of a finite number of
affine subspaces of the feature space. The affine subspaces have a common dimension, which is assumed to be much less than the dimension of the feature space. The subspaces are found using regression based on the $\ell_0$-norm. The proposed method is a generalisation of classical NN (Nearest Neighbor), NFL (Nearest Feature Line) classifiers and has a close relationship to NS (Nearest Subspace) classifier. The proposed classifier with an accurately estimated dimension parameter generally outperforms its competitors in terms of classification accuracy. We also propose a fast version of the classifier using a neighborhood representation to reduce its computational complexity. Experiments on publicly available datasets corroborate these claims.

~\\
\textsc{keywords}.~~
Intrinsic dimension estimation, Nearest constrained subspace classifier, Image classification, Sparse representation
\end{abstract}

\maketitle

\section{Introduction}
\label{section-Introduction}
The concept of data manifold
plays
a vital  role in pattern recognition.
Briefly speaking,
a data manifold is a topological space which contains the data samples, and
which serves as
an ideal geometric description of the data.
In this description, all data points, including the observed and unobserved, lie in a data manifold, whose
dimension is often much lower than the dimension of the feature space which contains it.

In previous work, the manifold model has been used as a powerful analytical approximation tool for nonparametric signal classes such as human face images or handwritten digits \cite{manifolds_handwritten_digits,Eigenfaces_recognition,Whitney_Reduction_Network}.
If the data manifold is learned,
then it can be exploited for
classifier design.
The manifold learning usually involves constructing a mapping from the feature space to a lower-dimensional space that is adapted to the training data and that
preserves the proximity of data points to each other.

There have been many works on manifold learning. For example, methods
such as ISOMAP (ISOmetric Mapping)~\cite{isomap2000}, Hessian Eigenmaps (also known as HLLE, Hessian Locally Linear Embedding) \cite{donoho2003hessian},
LLE (Local Linear Embedding)~\cite{LLE2000}, Maximum Variance Unfolding (MVU) \cite{weinberger2004unsupervised}, Local Tangent Space Alignment (LTSA) \cite{zhang2004principal} and Laplacian Eigenmap~\cite{Belkin01laplacianeigenmaps} have been introduced.
These methods learn a low-dimensional manifold under the constraint that the proximity properties of the nearby data
are preserved.

We propose a novel supervised classifier framework in which each class is modeled by
a union of \textcolor{black}{a finite number of} affine subspaces.
The proposed algorithm is superior to the traditional classifiers such as NN (Nearest Neighbor), NFL (Nearest Feature Line, proposed by S. Z. Li~\cite{Li99facerecognition}), NS (Nearest Subspace), etc.,
because the use of finite affine subspaces allows a more accurate description of the distribution of the data.

The reminder of this paper is organized as follows. In Section \ref{Background}, some background and related works about the classical classifiers including NN (Nearest Neighbor), NFL (Nearest Feature Line) and NS (Nearest Subspace) are briefly revisited.
In Section \ref{NM},
the classification model of NM (Nearest Manifold) and some classifier design principles are presented. Then, a novel constrained subspace framework named NCSC
and its fast version are proposed in Section \ref{section:NCSC}. Section \ref{section:experiment} gives the experimental results on several publicly available datasets.
In Section \ref{section:conclusion}, some concluding remarks are given.

\section{Background and Related Works}\label{Background}
We argue that the NN, NFL and NS classifiers can be incorporated into a unified framework. Before the detailed discussion, let's first briefly
revisit the theoretical background.

\subsection{NN and NFL}
The NN, NFL and NS classifiers base the classification of a sample $\mathbf{y}$ on the distances measured in the feature space.

\begin{figure}[htb]
  \centering
  \subfigure[NN]{
    \label{fig:Fig1-1}
    \includegraphics[height = 4.5cm]{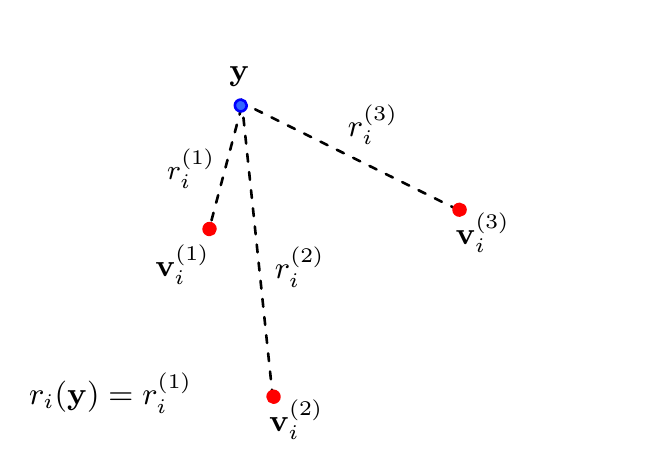}}
    \subfigure[NFL]{
    \label{fig:Fig1-2}
    \includegraphics[height = 4.5cm]{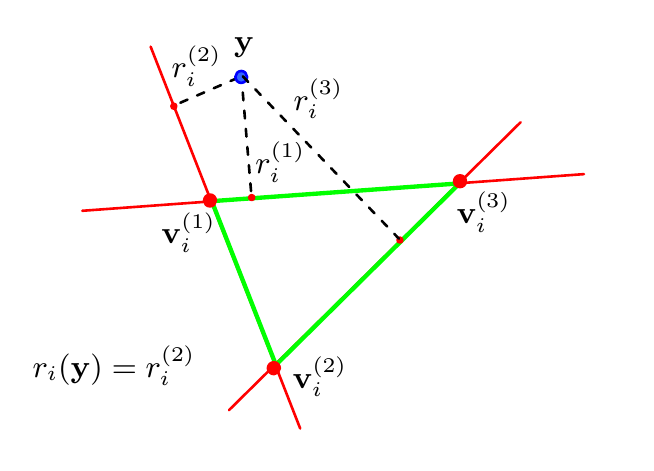}}
  \caption{The distances of a query sample $\mathbf{y}$ in NN and NFL to a class (the $i$-th, as shown), where $N_i = 3$. }
  \label{fig:Fig1}
\end{figure}

For NN and NFL, there
exists a convenient geometrical interpretation ---
given $N_i$ training samples in a given
class (say, the $i$-th class), the distances are obtained as illustrated in
Figure \ref{fig:Fig1} where to simplify the explanations,
we set $N_i = 3$.

In Figure \ref{fig:Fig1-1},
the distance from $\mathbf{y}$ to the $i$-th class is the minimum $r_i(\mathbf{y})$ of the distances from $\mathbf{y}$ to the training samples in the $i$-th class. In Figure \ref{fig:Fig1-2}, each pair of training samples defines a line. The distance from $\mathbf{y}$ to the $i$-th class is defined as the minimum $r_i(\mathbf{y})$ of the distances
from $\mathbf{y}$ to the different lines.

More generally, given the training samples $\mathbf{v}_i^{(1)}, \cdots,\mathbf{v}_i^{(N_i)}$ of the $i$-th class, in NN, $r_i(\mathbf{y})$ is written as
\begin{eqnarray}
\label{eq:NN}
r_i(\mathbf{y}) =
\mathop {\min }\nolimits_{j \in \{1,\cdots,N_i\}} {\left\| \mathbf{y} - \mathbf{v}_i^{(j)} \right\|_{2}}.
\end{eqnarray}
In NFL, $r_i(\mathbf{y})$ is defined as
\begin{eqnarray}
\label{eq:NFL}
r_i(\mathbf{y}) =
\mathop {\min }\nolimits_{\lambda \in \mathbbm{R}, a, b  \in \{1,\cdots,N_i\}} {\left\| \mathbf{y} - \lambda \mathbf{v}_i^{(a)} - (1 - \lambda )\mathbf{v}_i^{(b)} \right\|_{2}}.
\end{eqnarray}

\subsection{NS}
In NS, the minimum distance $r_i(\mathbf{y})$
is the projection distance from $\mathbf{y}$ to the subspace linearly spanned by
all the training samples in the $i$-th class.

More specifically, given $N_i$ training samples
in the $i$-th class, define
\begin{equation}
\label{eq:definitionV}
{\mathbf{V}_i} = \left[ {\mathbf{v}_i^{(1)},\cdots,\mathbf{v}_i^{({N_i})}} \right]
\end{equation}
where ${\mathbf{v}_i^{(j)}\in {\mathbbm{R}^D}}$ is the $j$-th training sample and $D$ is the feature dimension.

Note that we assume that the training samples ${\mathbf{v}_i^{(1)},\cdots,\mathbf{v}_i^{({N_i})}}$ are linearly independent. Namely,
$\mathbf{V}_i \in \mathbbm{R}^{D \times N_i}$ is a full rank matrix, satisfying $D \geqslant N_i$.
Henceforth, unless otherwise stated, we assume that the given training samples are
\textcolor{black}{linearly} independent. This assumption is satisfied in many pattern recognition problems in which the feature space has a high dimension.

Then, in NS,
$r_i(\mathbf{y})$
is defined as
\begin{equation}
\label{eq:NS-distance}
{r_i}(\mathbf{y}) = \mathop {\min }\nolimits_{{\boldsymbol{\alpha}_i} \in \mathbbm{R}^{N_i}} {\left\| {\mathbf{y} - {\mathbf{V}_i}{\boldsymbol{\alpha}_i}} \right\|_{2}}
\end{equation}
where ${\boldsymbol{\alpha}_i} \doteq {\left[ {\alpha _i^{(1)},\cdots,\alpha _i^{({N_i})}} \right]^\mathrm{T}}$ is the coefficient vector.

If the $N_i$ training samples are linearly independent, the spanned subspace is $N_i$-dimensional.

After the distances from $\mathbf{y}$ to $K$ classes are obtained,
NN, NFL and NS, using the same scheme, determine the class of $\mathbf{y}$ by
\begin{equation}
\label{eq:classOfY}
\operatorname{class}
(\mathbf{y}) = \mathop{\operatorname{argmin}}\nolimits_{i \in \{1,\cdots,K\}}{~r_i}(\mathbf{y}).
\end{equation}

\section{NM: Nearest Manifold}\label{NM}
\label{section:NM}
The NM (Nearest Manifold) classifier is a generalization of the NN, NFL and NS classifiers. It also determines the class of a query sample based on the minimum distance.
In NM,  the manifold associated with a given class is a topological space such that all the data points (including the observed ones and unobserved ones) of
the class are lying on or near to it.
Thus, the manifold dimension is actually the intrinsic dimension of the dataset and is usually much less than the dimension of feature space.
If a query data sample is near to a data manifold, then it is assigned to the corresponding class.

\subsection{Model}
For a given a class, we define its universal dataset to be the ``conceptual'' set
containing all the observed and unobserved data of this class. It is assumed that this universal data
set forms a manifold in the feature space, and that the dimension of the manifold is much less than the dimension of the feature space.

Given $K$ data manifolds denoted by $\mathcal{M}_1,\cdots,\mathcal{M}_K$,
$\mathbf{y}$ is assigned to the class whose data manifold is the nearest to
$\mathbf{y}$.
More specifically, ${r_i}(\mathbf{y})$ is written as follows.
\begin{eqnarray}
{r_i}(\mathbf{y}) = \mathop {\min }\nolimits_{{\mathbf{z}} \in \mathcal{M}_i} {\left\| {\mathbf{y} - {\mathbf{z}}} \right\|_{2}},\text{ }\forall i = 1,\cdots,K.
\end{eqnarray}

After that, NMC (Nearest Manifold Classifier) uses Equation (\ref{eq:classOfY}) to classify  ${\mathbf{y}}$.

\subsection{Classifier Design Based on the Least Distance}
\label{design-clue}
Although it is difficult to implement the NM classifier
primarily due to the difficulty of deducing the $K$ data manifolds from the given training samples,
the NM model gives us some clues for designing a good classifier based on the nearest distance.

Given the training sets for each of $K$ classes,
a good classifier can be obtained by adding new derived points to the training sets.
In NFL, the derived points consist of
points on the feature lines of the class.
In NS, the derived points consist of the subspaces
spanned by the training samples of the class.
The solution of NM is to use data manifolds $\mathcal{M}_1, \cdots, \mathcal{M}_K$ to replace the training sets.

From this point of view, NFL, and NS are approximations to NM
in that the training sets and the derived points approximate the data manifolds.
But these approximations are not necessarily the best.

\section{Nearest Constrained Subspace Classifier}\label{section:NCSC}
\label{section:Nearest constrained subspace classifier}
We propose a novel classifier called the nearest constrained subspace classifier (NCSC), which
generalizes NN, NFL and \textcolor{black}{has a close relationship to NS}. The proposed classifier is formulated as the solution to
a problem of constrained least-squares regression, i.e. $\ell_2$-norm minimization.

\subsection{Coefficient Constraints}
Given $N_i$ training samples ${\mathbf{v}_i^{(1)},\cdots,\mathbf{v}_i^{({N_i})}}$
in the $i$-th class ($\forall i = 1,\cdots,K$), NCSC defines $r_i(\mathbf{y})$ as
\begin{eqnarray}
\label{eq:constraint-normalization}
&r_i(\mathbf{y}) = \mathop {\min }\nolimits_{{\boldsymbol{\alpha} _i} \in {\mathbbm{R}^{{N_i}}}} {\left\| {\mathbf{y} - {\mathbf{V}_i}{\boldsymbol{\alpha} _i}} \right\|_2}& \nonumber\\
&{\text{ subject to }}\sum\nolimits_{j = 1}^{{N_i}} {{\alpha }_i^{(j)}}  = 1 \text{ and } {{\left\| \boldsymbol{\alpha}_i \right\|}_{0}} \leqslant \kappa \leqslant N_i&
\end{eqnarray}
where ${\alpha}_i^{(j)}$ is the $j$-th entry of $\boldsymbol{\alpha}_i$
and $\mathbf{V}_i$ is of full rank and defined in Equation (\ref{eq:definitionV}).

If the columns in $\mathbf{V}_i$ and $\mathbf{y}$ are linearly independent, then the constraint $\sum\nolimits_{j = 1}^{{N_i}} {{\alpha }_i^{(j)}}  = 1$
ensures that the affine space spanned by $\mathbf{V}_i\boldsymbol{\alpha}_i$ as $\boldsymbol{\alpha}_i$ varies does not include the origin.

Since the $\ell_0$-norm of $\boldsymbol{\alpha}_i$ is equal to the number of nonzero entries in $\boldsymbol{\alpha}_i$, the inequality
${{\left\| \boldsymbol{\alpha}_i \right\|}_{0}} \leqslant \kappa \leqslant N_i$ ensures that there are at most $\kappa$ columns in $\mathbf{V}_i$
which contribute to $\mathbf{V}_i\boldsymbol{\alpha}_i$.
Since $\boldsymbol{\alpha}_i$ has $N_i$ entries, there are $\binom {N_i} \kappa$ $\kappa$-combinations of the valid columns (training samples) in $\mathbf{V}_i$.
Note that as $\binom {N_i}  \kappa$ can be very large, we use
a strategy described in Section \ref{sec:fast-NCSC}, for reducing the number of $\kappa$-combinations of columns that are considered.

For the $m$-th training sample combination of the $i$-th class, we define the base matrix as follows.
\begin{eqnarray}
\label{eq:W}
&{\mathbf{W}_{i,m}} \doteq
\left[{\mathbf{w}_i^{(1)}},\cdots,{\mathbf{w}_i^{(\kappa)}}\right] &\nonumber \\
&~~\text{ subject to }~~ \left\{\mathbf{w}_i^{(1)},\cdots,\mathbf{w}_i^{(\kappa)}\right\} \subseteq \left\{{\mathbf{v}_i^{(1)}},\cdots,{\mathbf{v}_i^{(N_i)}}\right\}.&
\end{eqnarray}

Using Equation (\ref{eq:W}), the subproblem of Equation (\ref{eq:constraint-normalization}) for the above mentioned $\kappa$-combination can be rewritten as follows.
\begin{eqnarray}
\label{eq:constraint-normalization-reduced}
{r_i^{(m)}(\mathbf{y})}  = \mathop {\operatorname{argmin} }\limits_{{{\boldsymbol{\beta}} } \in {\mathbbm{R}^{{\kappa}}}} {\left\| {\mathbf{y} - {{\mathbf{W}_{i,m}}}{{\boldsymbol{\beta}} }} \right\|_2}
{\text{~~subject to }} &\sum\nolimits_{j = 1}^{{\kappa}} {{\beta }^{(j)}}  = 1&
\end{eqnarray}
where ${{\beta }^{(j)}}$ is the $j$-th entry of $\boldsymbol{\beta}$.

Note that there are very mature algorithms for Equation (\ref{eq:constraint-normalization-reduced}). Interested readers are referred to
\cite{Solve_constrained001,Solve_constrained002} for more details.

Using Equation (\ref{eq:constraint-normalization-reduced}), the solution of Equation (\ref{eq:constraint-normalization})
is obtained from
the $\binom {N_i}  \kappa$
sub-solutions in that
the distance $r_i(\mathbf{y})$ is defined by
\begin{eqnarray}\label{eq:NCSC-distance}
r_i(\mathbf{y}) = \min\limits_{m} r_i^{(m)}(\mathbf{y}).
\end{eqnarray}

\subsection{NCSC}
Based on Equation
(\ref{eq:constraint-normalization}), our proposed \emph{Nearest Constrained Subspace Classifier} (NCSC) is
summarized as Algorithm \ref{alg:NCSC}.

\begin{algorithm}[H]
\caption{Nearest Constrained Subspace Classifier}
\label{alg:NCSC}
  \SetAlgoLined
  \SetKwInOut{Input}{Input}
  \SetKwInOut{Output}{Output}
  \Input{A query sample $\mathbf{y}$, training vectors partitioned to $K$ classes and parameter $\kappa$.}
  \Output{Class ID of $\mathbf{y}$.}
  \BlankLine
  %\BlankLine
  %Define ${{\mathbf{V}_1},\cdots,{\mathbf{V}_K}}$ as in Equation (\ref{eq:definitionV})\;
  \For{$i$ $\leftarrow$ $1$ \KwTo $K$}
  {
      \For{$m$ $\leftarrow$ $1$ \KwTo $\binom {N_i}  \kappa$}
      {
        Obtain $\mathbf{W}_{i,m}$ as in Equation (\ref{eq:W})\;
        Calculate ${r_i^{(m)}(\mathbf{y})}$ as in Equation (\ref{eq:constraint-normalization-reduced})\;
      }

      Calculate ${r_i(\mathbf{y})}$ as in Equation (\ref{eq:NCSC-distance})\;

  }
  \Return $\operatorname{class}(\boldsymbol{y}) \leftarrow \mathop{\operatorname{argmin}}\nolimits_{i}{r_i}(\boldsymbol{y})$\;
\end{algorithm}

Note, when $\kappa = 1$, Equation (\ref{eq:constraint-normalization}) becomes Equation (\ref{eq:NN}) and when $\kappa = 2$, Equation (\ref{eq:constraint-normalization})
becomes Equation (\ref{eq:NFL}).
This shows that NN and NFL are just two special cases of NCSC.
Furthermore, we observe  that the training sets used in NN are zero-dimensional affine
subspaces, and the  sets of feature lines used in NFL
are one-dimensional affine subspaces.

If $\kappa = 3$, the NCSC becomes the NFP (Nearest Feature Plane) method.
More mathematically, for the NFP method,
when $N_i \geqslant 3$, Equation (\ref{eq:constraint-normalization})
can also be rewritten as follows.
\begin{eqnarray}
\label{eq:NFP}
&r_i(\mathbf{y}) =
\mathop {\min }\nolimits_{\lambda_1, \lambda_2 \in \mathbbm{R}} {\left\| \mathbf{y} - \lambda_1 \mathbf{v}_i^{(a)} - \lambda_2 \mathbf{v}_i^{(b)}
- (1 - \lambda_1 - \lambda_2)\mathbf{v}_i^{(c)} \right\|_{2}}& \nonumber\\
&\text{subject to}~ \left\{a, b, c \right\} \subseteq \left\{1,\cdots, N_i \right\}.&
\end{eqnarray}

In Equation (\ref{eq:NFP}), when $\lambda_1 > 0, \lambda_2 > 0$ and $1 - \lambda_1 - \lambda_2 > 0$,
the corresponding points are
inside the triangle determined by $\mathbf{v}_i^{(a)}, \mathbf{v}_i^{(b)}$ and $\mathbf{v}_i^{(c)}$,
otherwise the subspace points are outside the triangle or on its edges.

In Algorithm \ref{alg:NCSC},
the dimension of the affine subspaces is $\kappa - 1$.
For convenience, we denote $D_\text{m}, D_\text{c}, D_\text{s}, D$, respectively as the dimensions of the data manifold,
the affine subspace, the subspace used in NS and the feature space.

The value of $D_\text{c}$ is
\begin{equation}
\label{eq:L0-norm-dimension}
D_\text{c} = \kappa - 1 \geqslant 0.
\end{equation}

In NN, $D_\text{c} = 0$ and in NFL, $D_\text{c} = 1$.
In NS, there is no constraint on the $D_\text{s}$, then, given $N_i$ \textcolor{black}{linearly} independent training samples of a class, the dimension of subspace is the largest,
namely,
\begin{eqnarray}
D_\text{s} = N_i.
\end{eqnarray}

\subsection{Union of Affine Hulls and Data Manifold Approximation}
Given $K$ classes, there are $K$ constrained subspaces in NCSC.
Each constrained subspace is a union of affine hulls. Here we give the explanation as follows.
The $i$-th constrained subspace $\mathbbm{S}_{i}^{\text{NCSC}} $  is written as follows.
\begin{equation}\label{eq:constrained-subspace-sparse-epresentation}
\mathbbm{S}_{i}^{\text{NCSC}} = \left\{\mathbf{V}_i \boldsymbol{\alpha} |~
\boldsymbol{\alpha} \in \mathbbm{R}^{N_i},
\|\boldsymbol{\alpha}\|_0 \leqslant \kappa \leqslant N_i
~\text{and}~ \mathbf{1}^\mathrm{T}\boldsymbol{\alpha} = 1
\right\}.
\end{equation}

Since the problem of Equation (\ref{eq:constraint-normalization}) can be divided into the $\binom {N_i} \kappa$  subproblems
of Equation (\ref{eq:constraint-normalization-reduced}), it is
not difficult to show that $\mathbbm{S}_{i}^{\text{NCSC}}$  can be rewritten as a union of affine hulls. Namely
\begin{equation}
\left\{
\begin{aligned}
&\mathbbm{S}_{i}^{\text{NCSC}} = \mathop{\textstyle\bigcup}\nolimits_{m=1}^{\binom {N_i} \kappa} \mathcal{H}_{i,m}\\
&\mathcal{H}_{i,m} = \left\{\mathbf{W}_{i,m} \boldsymbol{\beta}|~\boldsymbol{\beta}
\in \mathbbm{R}^{\kappa}~{\text{and}}~\mathbf{1}^\mathrm{T}\boldsymbol{\beta} = 1 \right\}
\end{aligned}
\right.
\end{equation}
where $\mathcal{H}_{i,m}$ is the affine hull of the column vectors in $\mathbf{W}_{i,m}$.

Note that affine hull (i.e., affine subspace) is also referred to as “linear manifold”. This means
that $\mathbbm{S}_{i}^{\text{NCSC}}$  can be viewed as an approximation to $\mathcal{M}_i$  by using a series of linear manifolds.

Let $\mathbbm{S}_{i}^{\text{NS}}$ denote
the linear subspace spanned by the $N_i$ training samples of the $i$-th class. $\mathbbm{S}_{i}^{\text{NS}}$ is  written as follows.
\begin{equation}\label{eq:subspace-formulation}
\mathbbm{S}_{i}^{\text{NS}} = \left\{\mathbf{V}_i \boldsymbol{\alpha} |~
\boldsymbol{\alpha} \in \mathbbm{R}^{N_i}
\right\}.
\end{equation}

From Equations (\ref{eq:constrained-subspace-sparse-epresentation}) and (\ref{eq:subspace-formulation}), it is not difficult to find
that $\mathbbm{S}_{i}^{\text{NCSC}}$ is a subset of $\mathbbm{S}_{i}^{\text{NS}}$
and in NCSC, the constrained subspaces defined as $\kappa$ increases are nested. Namely
\begin{eqnarray}\label{eq:nest}
\mathbbm{S}_{i}^{\text{NCSC}}|_{\kappa = 1} \subset \mathbbm{S}_{i}^{\text{NCSC}}|_{\kappa = 2} \subset \cdots \subset \mathbbm{S}_{i}^{\text{NCSC}}|_{\kappa = N_i}
\subset \mathbbm{S}_{i}^{\text{NS}}.
\end{eqnarray}

Note that $\mathbbm{S}_{i}^{\text{NCSC}}|_{\kappa = 1}, \cdots, \mathbbm{S}_{i}^{\text{NCSC}}|_{\kappa = N_i}$
and $\mathbbm{S}_{i}^{\text{NS}}$ can be regarded as some approximations to the data manifold
$\mathcal{M}_i$.
In this sense, we argue that NCSC/NS  is an approximation to NMC (Nearest Manifold Classifier)
and among $\mathbbm{S}_{i}^{\text{NCSC}}|_{\kappa = 1}, \cdots, \mathbbm{S}_{i}^{\text{NCSC}}|_{\kappa = N_i},
\mathbbm{S}_{i}^{\text{NS}}$, only one is the best approximation to
$\mathcal{M}_i$.
The approximation degree
can be measured by the classification accuracies of NCSC with a variety of $\kappa$ (or equivalently $D_\text{c}$) and NS.

In order to make $\mathbbm{S}_{i}^{\text{NCSC}}$ the best approximation to $\mathcal{M}_i$,
we argue that the common dimension of the affine subspaces in $\mathbbm{S}_{i}^{\text{NCSC}}$ should be equal to the dimension of $\mathcal{M}_i$.
Otherwise the effectiveness of the linear approximation to a nonlinear manifold can not be guaranteed.  Namely, we have
\begin{equation}
\label{eq:dimension-equation}
{D_\text{c}} = {D_\text{m}}.
\end{equation}

In many applications,
$D$ is far larger than $D_\text{s}$, $D_\text{c}$ and $D_\text{m}$, namely,
\begin{equation}
\label{eq:dimension-inequality}
D_\text{m} = D_\text{c} < D_\text{s} = N_i\ll D.
\end{equation}

Although there are lots of techniques of feature dimension reduction for improving the algorithm efficiency,
we argue that if the feature dimension $D$ is near to $D_\text{s}$, or equivalently, $D$ is near to $N_i$ (the $N_i$ training
samples of the $i$-th class are \textcolor{black}{linearly} independent),
the effectiveness of NS can not be guaranteed
because
$r_1(\mathbf{y}),\cdots, r_K(\mathbf{y})$
are all near to $0$.

Also of particular note in Equation (\ref{eq:dimension-inequality}) is that, for a guaranteed performance of NCSC (and other classifiers), there exists a lower
bound on $N_i$, which is also crucial for estimating $D_\text{m}$. More specifically, the lower bound should not be less than $D_\text{m}$.
Namely, we have
\begin{equation}
\label{eq:sample-number-inequality}
N_i > D_\text{m}, \,\,\forall i = 1,\cdots,K.
\end{equation}
Otherwise, the training samples are insufficient to guarantee an accurate estimate of $D_\text{m}$ and high classification accuracy.

\subsection{Computational Complexity and Fast NCSC}
\label{sec:fast-NCSC}
Given $N_i$ training samples in the $i$-th class, the problem of Equation (\ref{eq:constraint-normalization}) is decomposed to $\binom {N_i} \kappa$
subproblems of Equation (\ref{eq:constraint-normalization-reduced}).
If $N_i$ is large and $\kappa \simeq  \frac{N_i}{2}$, then ${\binom {N_i} \kappa}$ can be huge.
This makes the computational
complexity of Algorithm \ref{alg:NCSC} very high.

To reduce the computational complexity of Algorithm \ref{alg:NCSC}, one simple strategy is to reduce
$\binom {N_i} \kappa$ to a smaller number.
To do this, we assume that the nearest $(\kappa - 1)$ neighbors of a data point $\mathbf{x}$ and the data point itself in the same class are sufficient to
capture the local manifold dimension at $\mathbf{x}$.
Note that the neighborhood strategy is also employed in \cite{levina2004maximum,carter2010local} to estimate the intrinsic dimension of a dataset under the same assumption.
We define the base matrix, whose columns vectors are $\mathbf{x}$ and its $(\kappa - 1)$ neighbors in the same class, as follows.
\begin{equation}
\label{eq:V-x}
\mathbf{V}(\mathbf{x}) \doteq \left[\mathbf{v}_1,\cdots,\mathbf{v}_{\kappa}\right]
\end{equation}
where $\mathbf{x} \in \{\mathbf{v}_1,\cdots,\mathbf{v}_{\kappa}\}$.

We implement fast NCSC, i.e., the NCSC via $\kappa$-neighbor representation, called NCSC-II, using Algorithm \ref{alg:fastNCSC}.

\begin{algorithm}[H]
\caption{NCSC-II --- fast NCSC via $\kappa$-neighbor representation}
\label{alg:fastNCSC}
  \SetAlgoLined
  \SetKwInOut{Input}{Input}
  \SetKwInOut{Output}{Output}
  \Input{A query sample $\mathbf{y}$, training vectors $\{\mathbf{x}_1,\cdots,\mathbf{x}_{N}\}$ partitioned to $K$ classes and parameter $\kappa$.}
  \Output{Class ID of $\mathbf{y}$.}
  \BlankLine
  %\BlankLine
  \For{$k$ $\leftarrow$ $1$ \KwTo $N$}
  {
      Calculate $\mathbf{V}(\mathbf{x}_k) $ as in Equation (\ref{eq:V-x})\;

      $r(\mathbf{y}, \mathbf{x}_k) \leftarrow  \mathop {\min }\nolimits_{{{\boldsymbol{\beta}} } \in {\mathbbm{R}^{{\kappa}}}} {\left\| {\mathbf{y} -
      {{\mathbf{V}(\mathbf{x}_k)}}{{\boldsymbol{\beta}} }} \right\|_2}
     {~~\text{  subject to  }}~~ \mathbf{1}^\mathrm{T}{\boldsymbol{\beta}}  = 1$\;

  }
  $m \leftarrow \mathop{\operatorname{argmin}}\nolimits_{k} r(\mathbf{y}, \mathbf{x}_k)$

  \Return $\operatorname{class}(\mathbf{y}) \leftarrow \operatorname{class}(\mathbf{x}_{m})$\;
\end{algorithm}

Given parameter $\kappa$  and training samples $\mathbf{x}_i^{(1)},\cdots,\mathbf{x}_i^{(N_i)}$  of the  $i$-th class,
$\mathbbm{S}_{i}^{\text{NCSC}}$ is formulated as the union of ${\binom {N_i} \kappa}$  affine hulls.
Since the complexity of Algorithm \ref{alg:NCSC} depends on the $\kappa$-combinations of training samples,
Algorithm \ref{alg:NCSC} is NP-hard.
In NCSC-II, most of the affine hulls are removed and the remaining ones define
$\mathbbm{S}_{i}^{\text{NCSC-II}}$,  based on Equation (\ref{eq:V-x}) as follows.
\begin{equation}
\left\{
\begin{aligned}
&\mathbbm{S}_{i}^{\text{NCSC-II}} = \mathop{\textstyle\bigcup}\nolimits_{m=1}^{N_i} \mathcal{H}_{i,m}\\
&\mathcal{H}_{i,m} = \left\{\mathbf{V}(\mathbf{x}_i^{(m)}) \boldsymbol{\beta}|~\boldsymbol{\beta}
\in \mathbbm{R}^{\kappa}~{\text{and}}~\mathbf{1}^\mathrm{T}\boldsymbol{\beta} = 1 \right\}
\end{aligned}
\right.
\end{equation}

By using the neighborhood representation, the computational complexity  of Algorithm \ref{alg:fastNCSC} is reduced to
$O(N)$.
We note that the properties/definitions of (\ref{eq:L0-norm-dimension}), (\ref{eq:nest}), (\ref{eq:dimension-equation}),
(\ref{eq:dimension-inequality}) and (\ref{eq:sample-number-inequality}) in NCSC still hold in NCSC-II.

\subsection{Intrinsic Dimension Estimation}
We argue that NCSC or NCSC-II with the fine-tuned parameter
is expected to be
the closest in performance to NMC (Nearest Manifold Classifier), which is believed to have the optimal classification accuracy. Furthermore, we contend
that the classification accuracy of NCSC/NCSC-II is a function of
$D_\text{c}$. We denote the function as $f(D_\text{c})$.

We now have a scheme for estimating $D_\text{m}$ of a labeled dataset.
$D_\text{m}$ is given on the assumption as follows.
\begin{equation}
\label{eq:manifold-dimension-estimate}
D_\text{m} = \mathop{\operatorname{argmax}}\limits_{D_\text{c}}{f(D_\text{c})}.
\end{equation}

Equation (\ref{eq:manifold-dimension-estimate}) gives rise to two observations. First, given a labeled dataset,
$D_\text{m}$ is estimated by NCSC/NSCS-II.
The second is that when $D_\text{m}$ is learned, we have a tuned
NCSC/NCSC-II, which outperforms many of its rivals such as NN, NFL, NS.

Note the NCSC/NCSC-II only has one parameter $D_\text{c}$ (or equivalently, $\kappa$). This assumes that all the classes have the same (local) intrinsic dimension.
We call a dataset homogeneous, if the data classes are of the same (local) dimension, and a dataset inhomogeneous, if the data classes are of different dimension.
For an inhomogeneous dataset, query samples can be classified by NCSC/NCSC-II if $D_\text{c}$ is equal to the average (rounded) intrinsic dimension.

\section{Experimental Results}\label{section:experiment}
In this section, we apply the NCSC/NCSC-II to several publicly available image datasets.
Note that the feature extraction usually serves as an important step for image classification. Effective features are beneficial to improve the classification accuracy.
But it is not the focus of this study, thus, we keep the feature extraction simple.

In the following experiments, for feature extraction, we subtract the mean from each vectorized image
and normalize the
feature vectors to have a unit $\ell_2$-norm. Vectorization is carried out by concatenating the
columns of each image.

\subsection{Evaluation of NCSC as Dimension Estimator}
\subsubsection{PICS Dataset}
First, we present the results on the PICS/PES dataset \cite{PICS_database}.
The PICS/PES dataset is relatively small and contains
$84$ cropped facial images belonging to $12$ subjects ($7$ images/subject $\times$ $12$ subjects)
with fixed eye location and different expressions.
The image size is $241 \times 181$ pixels.

\begin{figure}[htbp]
  \centering
  \begin{tabular}{c}
    \includegraphics[width=0.1\textwidth] {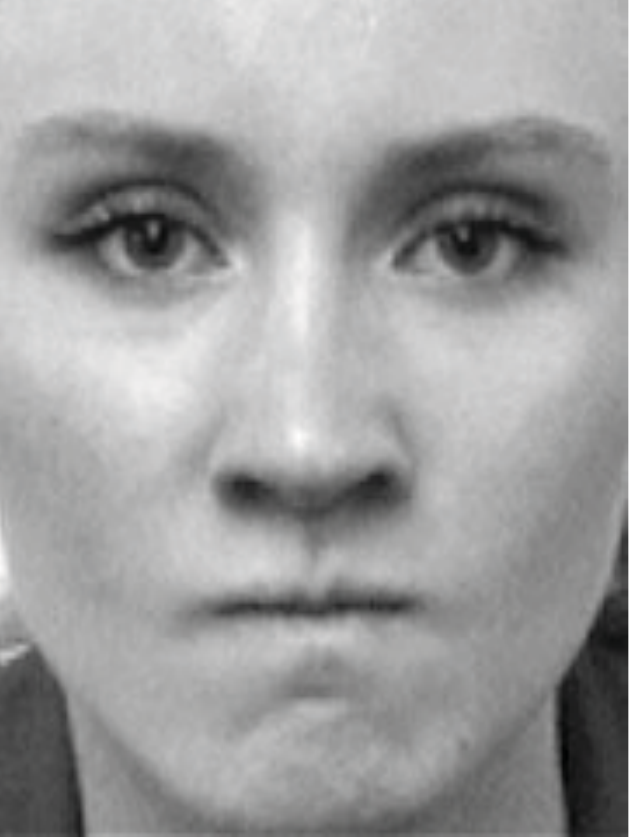}
    \includegraphics[width=0.1\textwidth] {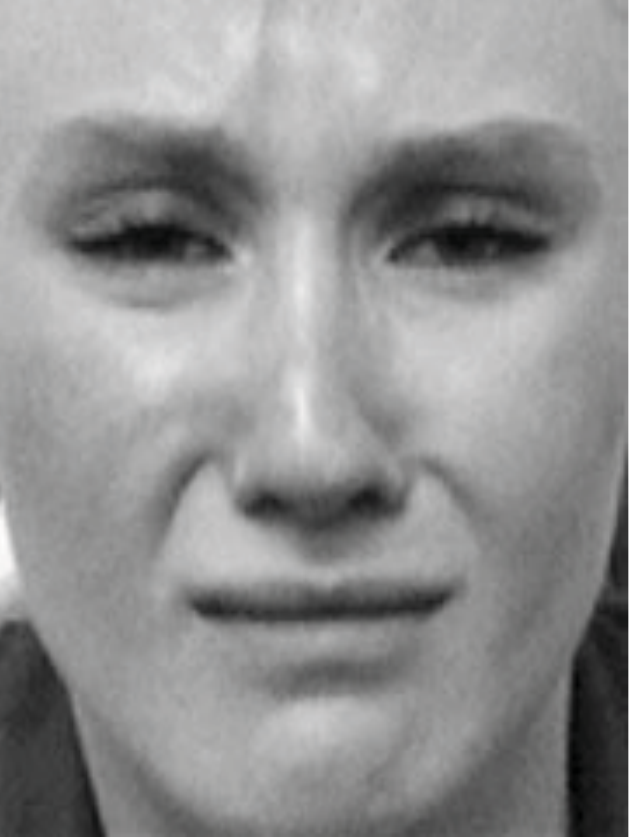}
    \includegraphics[width=0.1\textwidth] {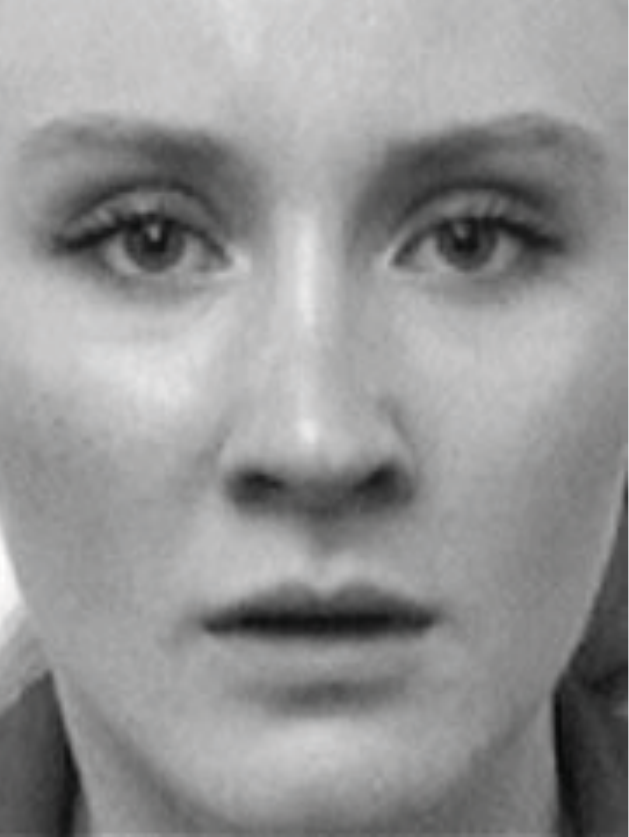}
    \includegraphics[width=0.1\textwidth] {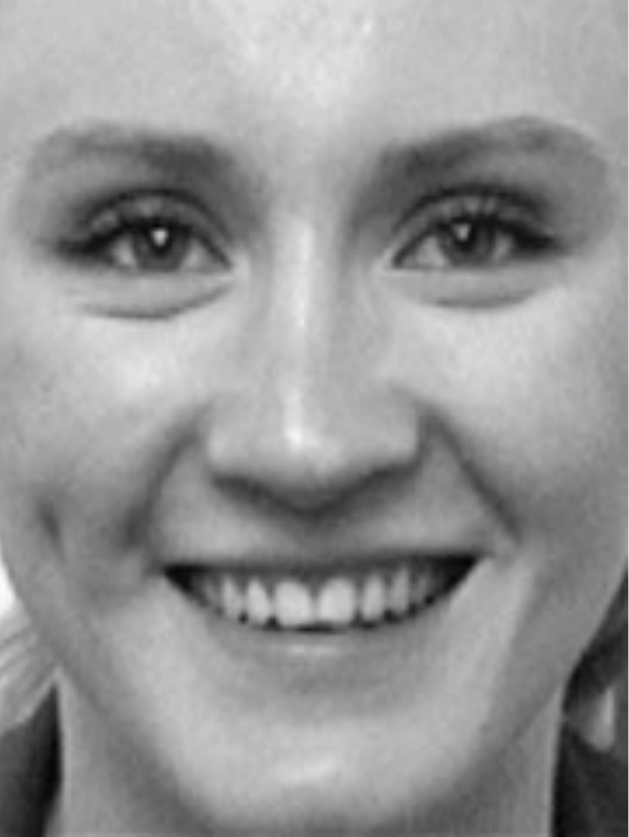}
    \includegraphics[width=0.1\textwidth] {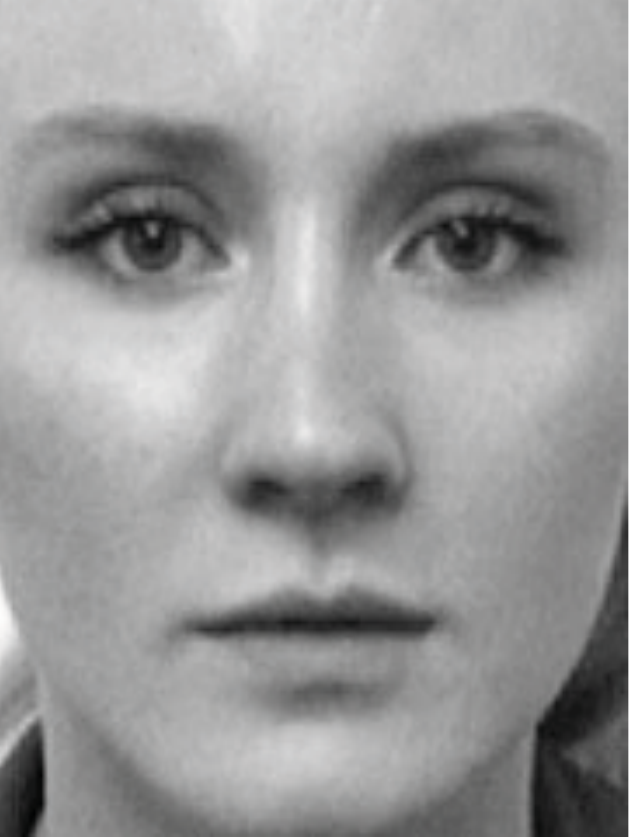}
    \includegraphics[width=0.1\textwidth] {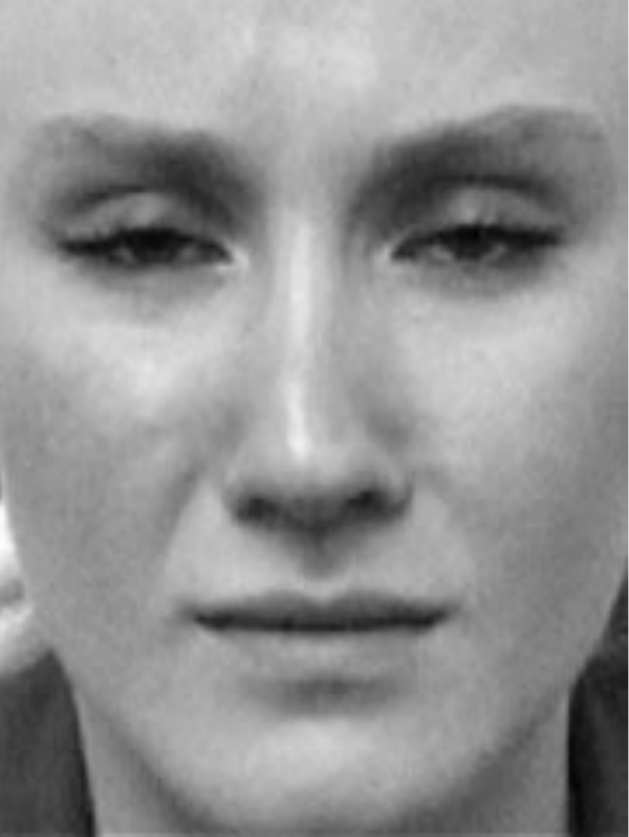}
    \includegraphics[width=0.1\textwidth] {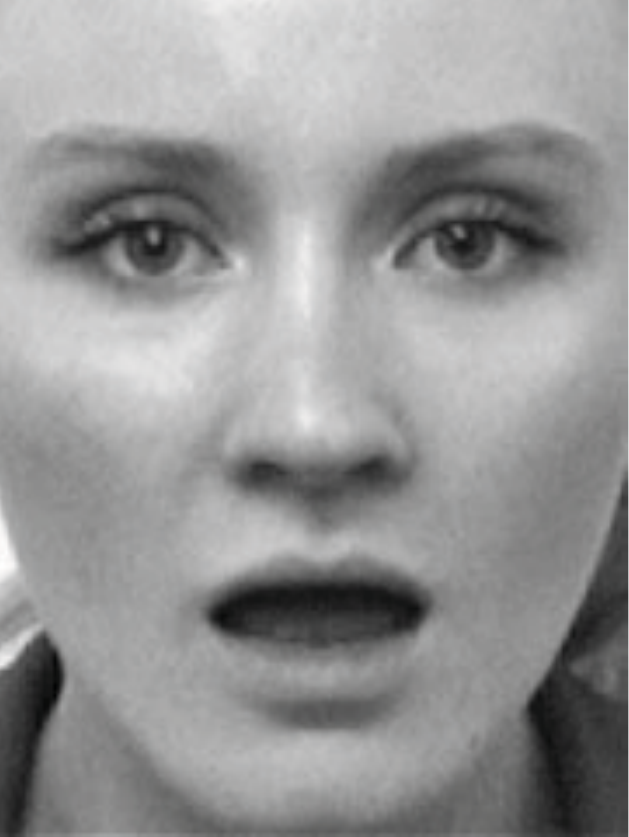}\\
    \includegraphics[width=0.1\textwidth] {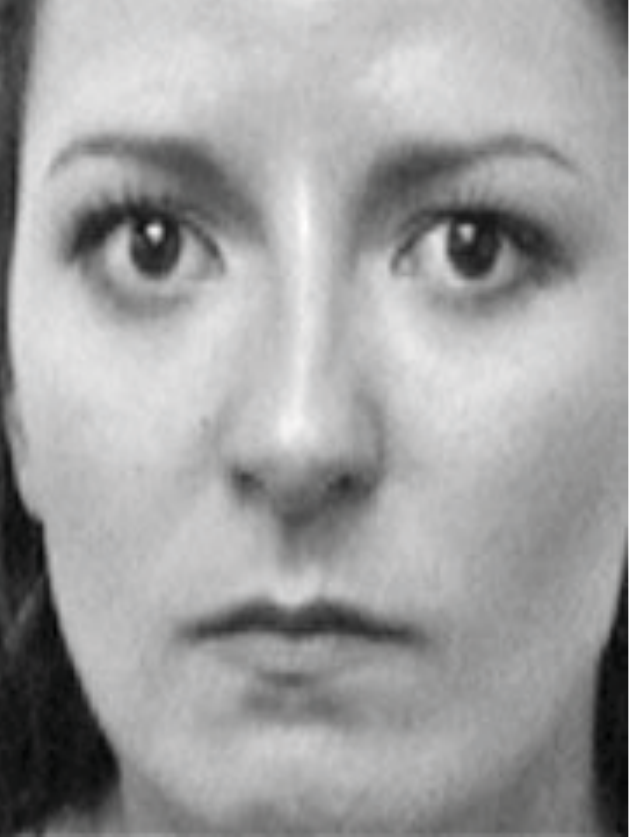}
    \includegraphics[width=0.1\textwidth] {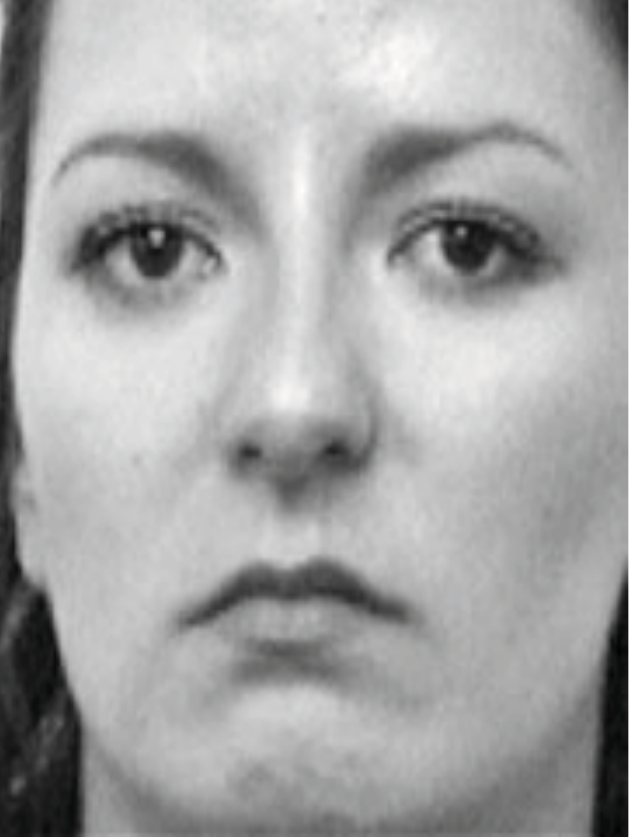}
    \includegraphics[width=0.1\textwidth] {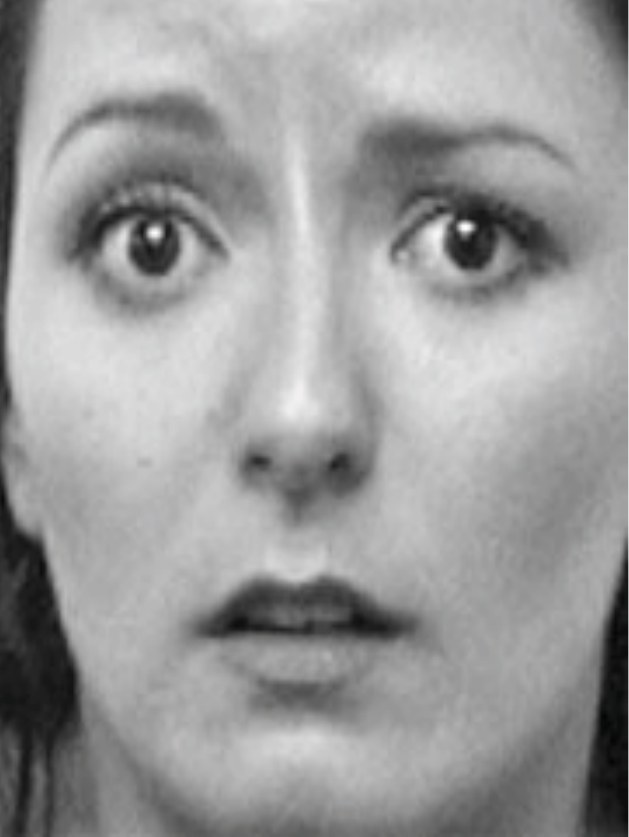}
    \includegraphics[width=0.1\textwidth] {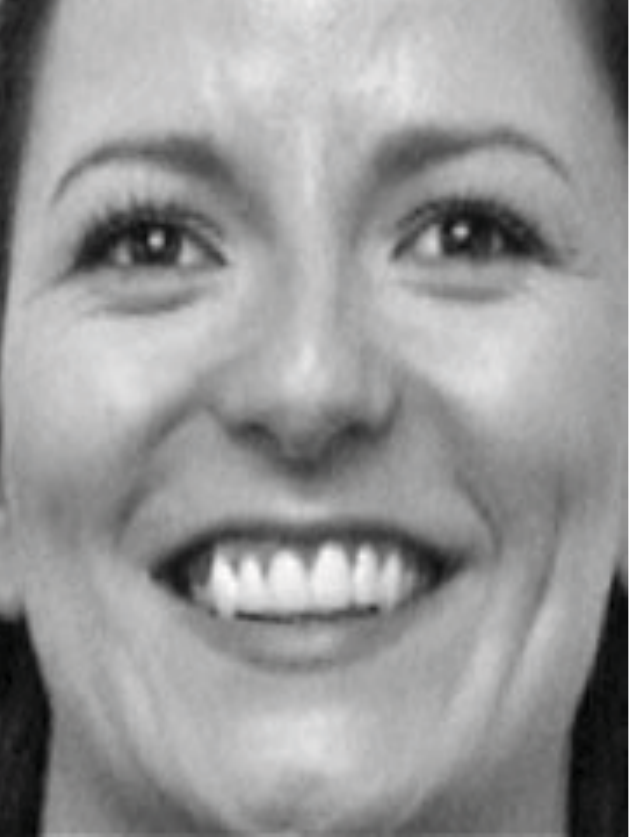}
    \includegraphics[width=0.1\textwidth] {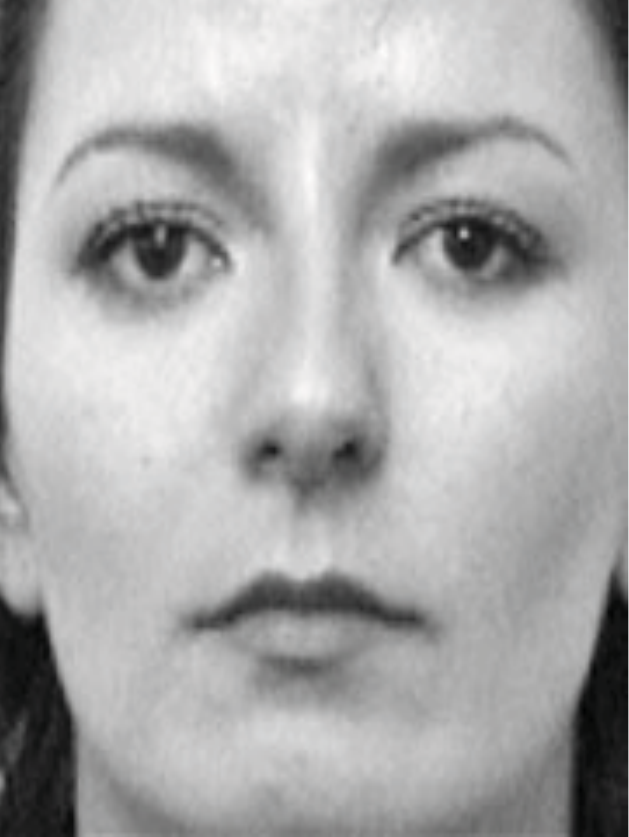}
    \includegraphics[width=0.1\textwidth] {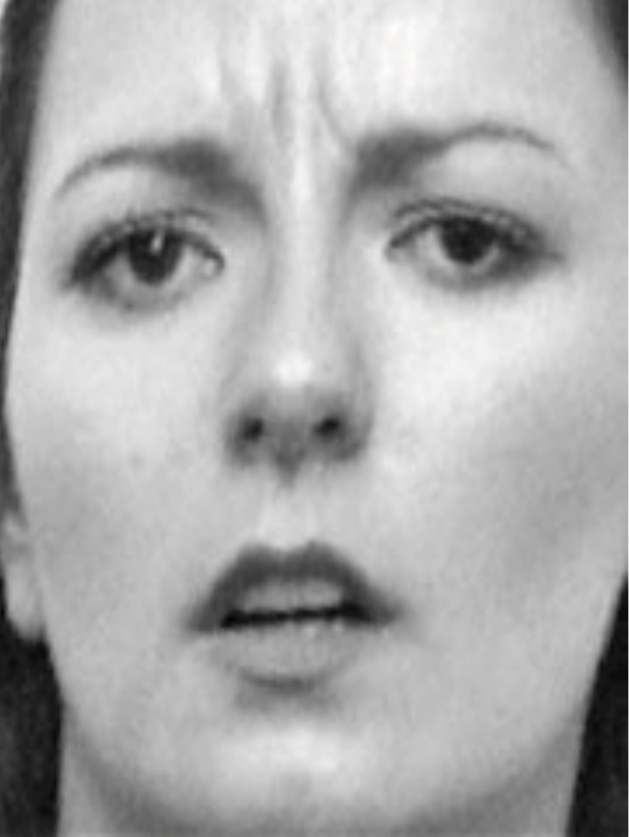}
    \includegraphics[width=0.1\textwidth] {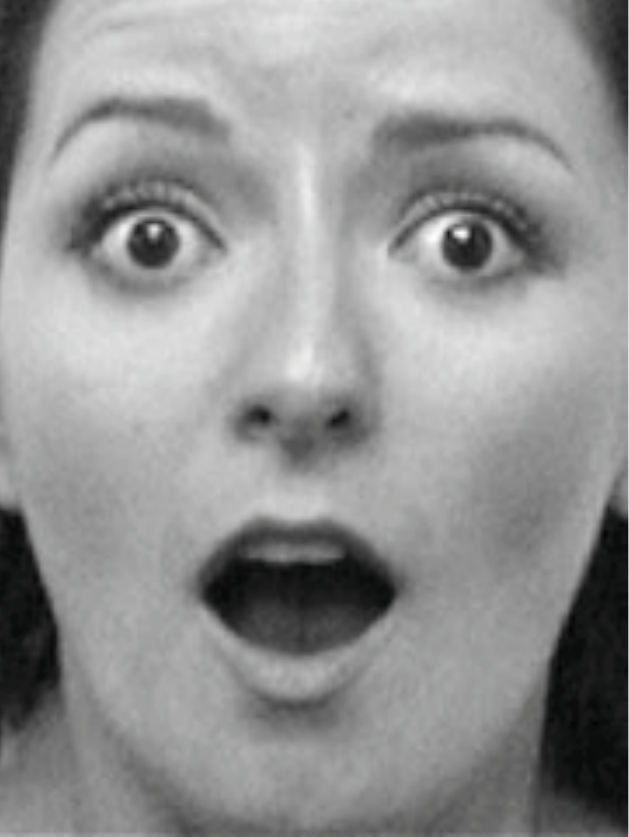}
  \end{tabular}
  \caption{Image examples of the PICS/PES dataset. First row: images of one subject.
  Second row: images of another subject.}
  \label{fig:pain-crops}
\end{figure}

Figure \ref{fig:pain-crops} shows the images of two subjects.

In order to get the classification accuracy from sufficient tests,
the experiment contains multiple rounds of classifications.
In each round of classification, we randomly select one image from each subject as the query samples (i.e., $1$ image/subject $\times$ $12$ subjects). The remaining $6$ images of the same subject  are chosen as the training samples (i.e., $6$ images/subject $\times$ $12$ subjects).

\begin{figure}[htb]
  \centering
  \begin{tabular}{c}
  \includegraphics[width=0.8\textwidth] {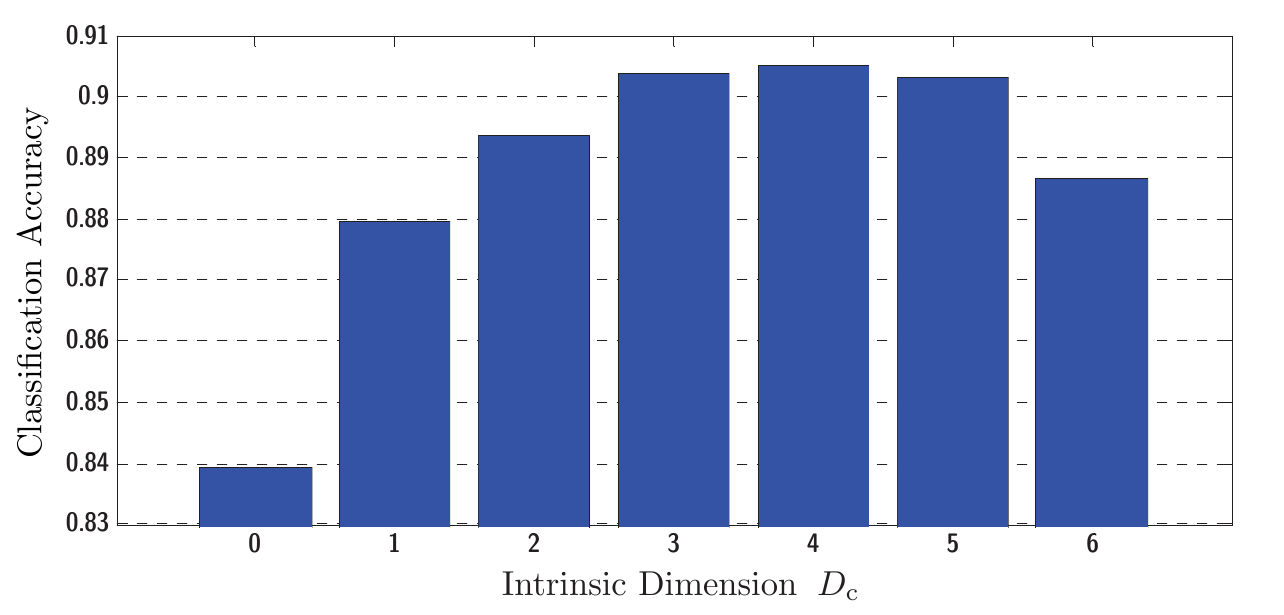}
  \end{tabular}
  \caption{Classification accuracy of NCSC on the PICS/PES dataset over varying $D_\text{c}$.}
  \label{fig:fig_pain_expression_classification}
%raw_data:     0.8393    0.8797    0.8937    0.9037    0.9050    0.9032    0.8867
\end{figure}

Figure \ref{fig:fig_pain_expression_classification} gives the classification rates of NCSC on the PICS/PES data with varying intrinsic dimension.
Each classification accuracy in Figure \ref{fig:fig_pain_expression_classification} is obtained as the classification accuracy of $2016$ query samples in $168$ rounds
($12$ query images/round $\times$ 168 rounds).

After one round is complete, we randomly select the query and training samples for another round of classifications.
After $168$ rounds, the classification accuracy is calculated.
The classification accuracy is given as follows.
\begin{equation}
\label{eq:classification-accuracy}
f(D_\text{c}) = \frac{w}{W}
\end{equation}
where $w$ is the number of the correctly classified query samples and $W$ is the total number of query samples. In this experiment, $W = 2016$.

Note that since the largest parameter $\kappa$ of NCSC on the $72$ training samples ($6$ training images/subject $\times$ $12$ subjects) is $6$ and $D_\text{c} = \kappa -1$,
the largest value of $D_\text{c}$ is $5$. In \textcolor{black}{order} to make a comprehensive comparison in Figure \ref{fig:fig_pain_expression_classification},
we also give the classification accuracy of NS. When $D_\text{c} = 6$, it actually means the result of NS with $D_\text{s} = 6$.

In order to avoid unnecessary perturbations in the accuracies, the same random selections
of training and query sets are preserved and repeated for different $D_\text{c}$.
It is clear that none of NN ($D_\text{c} = 0$), NFL ($D_\text{c} = 1$) and NS ($D_\text{c} = 6$)
achieves the optimal classification.

According to the highest classification accuracies in Figure \ref{fig:fig_pain_expression_classification},
our estimate of the intrinsic dimension of the PICS/PES dataset is $D_\text{m} = 4$.

\subsubsection{ORL Dataset}
Second, we give the experimental results on the ORL
face dataset \cite{orl_database}.
The dataset contains $400$ images ($10$ images/subject $\times$ $40$ subjects).
The image size is $112 \times 92$ pixels.
The dimension of the data manifold of each subject is believed to be larger than one
since the images were taken  at different times, with variations in the lighting, facial expressions, facial details and poses.

\begin{figure}[htb]
  \centering
  \begin{tabular}{c}
    \includegraphics[width=0.12\textwidth]{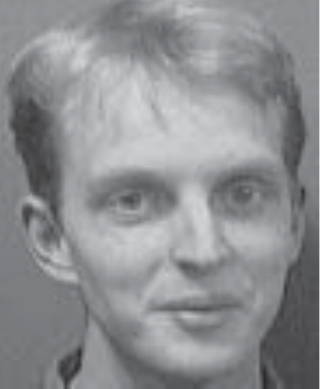}
    \includegraphics[width=0.12\textwidth]{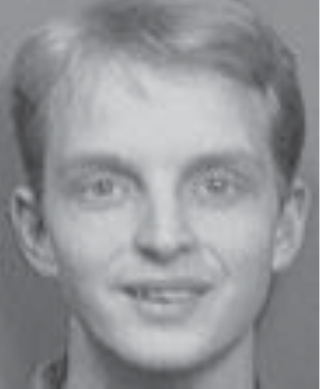}
    \includegraphics[width=0.12\textwidth]{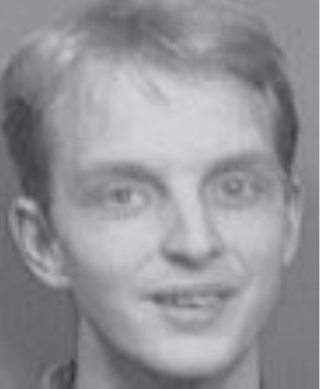}
    \includegraphics[width=0.12\textwidth]{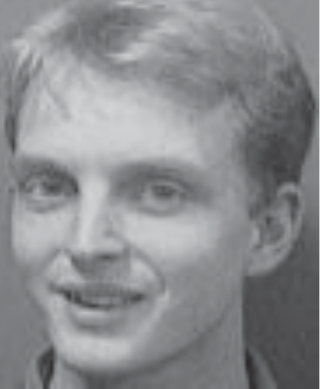}
    \includegraphics[width=0.12\textwidth]{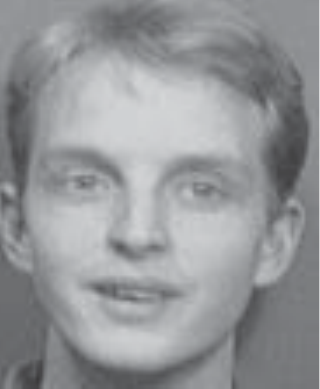}\\
    \includegraphics[width=0.12\textwidth]{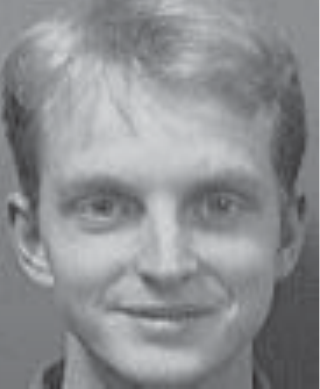}
    \includegraphics[width=0.12\textwidth]{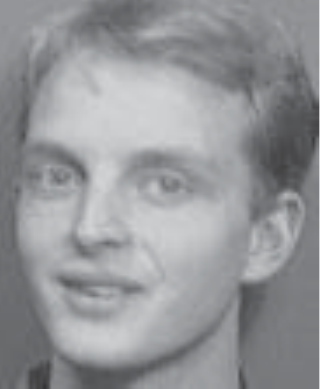}
    \includegraphics[width=0.12\textwidth]{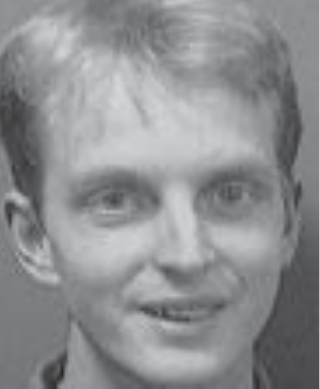}
    \includegraphics[width=0.12\textwidth]{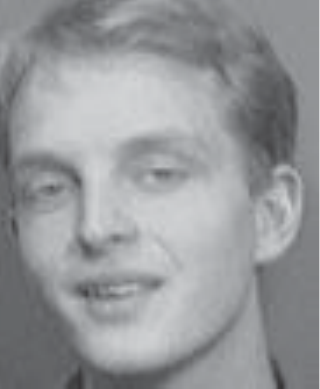}
    \includegraphics[width=0.12\textwidth]{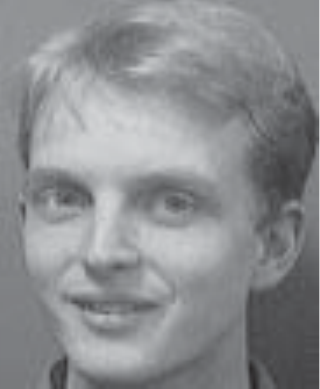}
  \end{tabular}
  \caption{Image examples of a subject, with varying imaging conditions, such as illumination, facial expressions, details and orientations.}
  \label{fig:ORL-example}
\end{figure}

Figure \ref{fig:ORL-example} shows the image examples of one subject from the ORL dataset.

Since the ORL dataset ($400$ images) is larger than the PICS/PES dataset ($84$ images), in order to avoid to the computational complexity problem of NCSC, we use NCSC-II for this experiment.

In order to be able to estimate  $D_\text{m}$, we set $N_i = 9$ for each class.
In this setting, in one round of classifications,
we have $360$ training images (i.e., $9$ training images/subject $\times$ $40$ subjects) and $40$  query images (i.e., $1$ query image/subject $\times$ $40$ subjects).
After one round is completed, we randomly select the query and training samples for another round.

The range of $D_\text{c} = 0,\cdots,9$ is traversed.
Note that when $D_\text{c} = 9$, the result is actually the accuracy of NS with $D_\text{s} = 9$.

\begin{figure}[htbp]
  \centering
    \includegraphics[width=0.8\textwidth]{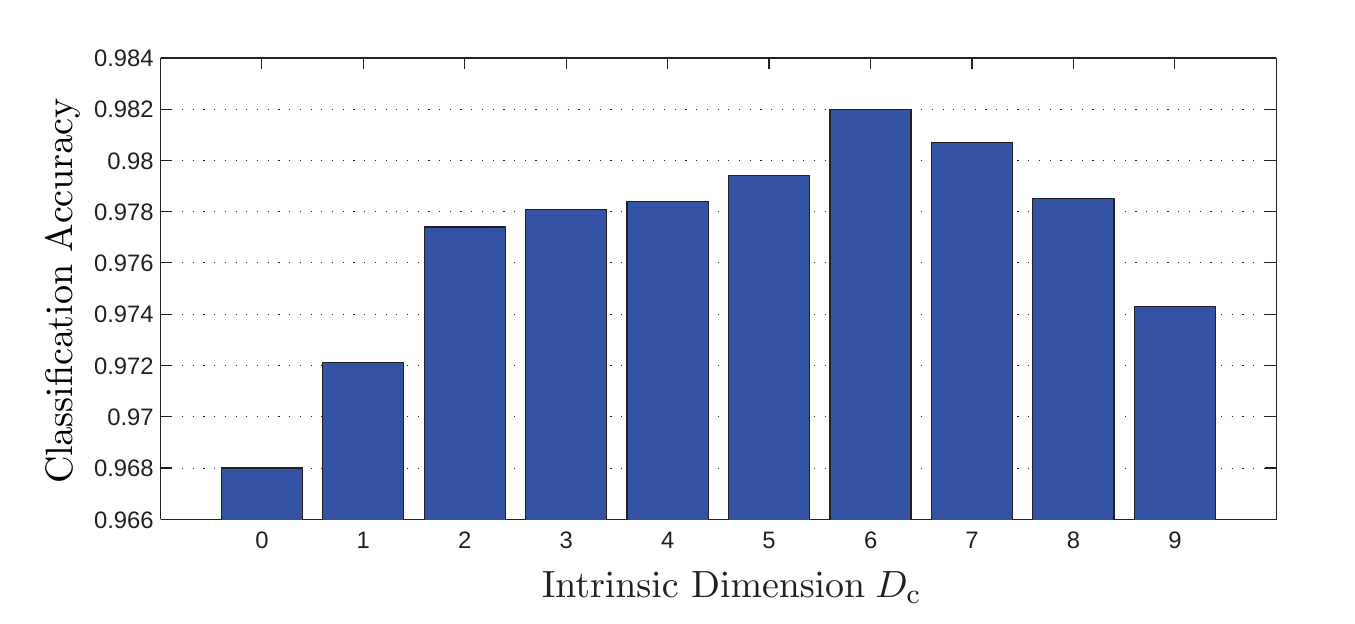}
    \caption{Classification accuracy on the ORL dataset for varying $D_\text{c}$.}
  \label{fig:Fig4}
  %raw:  0.9675    0.9745    0.9738    0.9750    0.9760    0.9766    0.9765    0.9760    0.9755    0.9743--------     NSCS
  %raw:  0.9680    0.9721    0.9774    0.9781    0.9784    0.9794    0.9820    0.9807    0.9785    0.9743--------fast NSCS
\end{figure}

Figure \ref{fig:Fig4} shows the classification accuracies of NCSC-II for the different $D_\text{c}$.
Each accuracy is obtained as the result of $8000$ query samples from $200$ classification rounds (i.e., $40$ query samples/round $\times$ $200$ rounds).
The classification accuracy is given by Equation
(\ref{eq:classification-accuracy})
with $W = 8000$.

As in the first classification experiment, the random selections
of training and query sets are preserved and repeated for classifications using different $D_\text{c}$.

Figure \ref{fig:Fig4} shows that the highest classification accuracy appears at $D_\text{c} = 6$.

\subsubsection{Comparison with Other Intrinsic Dimension Estimators}
Some well-known dimension estimators
include MLE (Maximum Likelihood Estimation) \cite{levina2004maximum}, Correlation
Dimension (hereafter referred to as Corr.Dim in this paper) \cite{grassberger1983measuring, camastra2002estimating}, PCA \cite{fukunaga1971algorithm,bruske1998intrinsic}
and their recent variations \cite{massoud2007manifold,fan2009intrinsic,carter2010local}, etc.
The statistical properties of these estimators, the best ways of evaluating them and the comprehensive comparisons of them are still open
research areas.

In this part,
we  make
a brief comparison between NCSC/NCSC-II and other estimators of PCA, MLE and Corr.Dim.

Among them, the PCA estimator, from its early proposal \cite{fukunaga1971algorithm}, to its later variants \cite{bruske1998intrinsic}, is based on
the classical principle component analysis, in which the estimate of dimension is determined by the number of eigenvalues not less than a predefined threshold.

The Corr.Dim estimator is summarized as follows. Given
a dataset $\left\{\mathbf{x}_1,\cdots, \mathbf{x}_{N_i}\right\}$, the following function is defined.
\begin{equation}
C(r) = \frac{2}{{N_i}({N_i}-1)}\sum\limits_{i=1}^{N_i}{\sum\limits_{j=i+1}^{N_i}{ H(r - \left\| {\mathbf{x}_{i}}-{\mathbf{x}_{j}} \right\|_2)}}
\end{equation}
where $H(\cdot)$ is a unit step function satisfying if $\upsilon > 0$, then
$H(\upsilon) = 1$, otherwise, $H(\upsilon) = 0$.
The intrinsic dimension is estimated by
plotting $\log C(r)$ against $\log r$ and calculating the slope of its linear part.
Interested readers are referred to \cite{grassberger1983measuring, camastra2002estimating} for more details.

The MLE estimator determines its estimate from $\{\mathbf{x}_1,\cdots, \mathbf{x}_{N_i}\}$ under the assumption
that the closest $k$ (where $k$ is a fixed number and $k > 2$) neighbors to a given point $\mathbf{x}_i$ lie on the same manifold.
The intrinsic dimension $\hat{C}_m$ is estimated as follows.

\begin{eqnarray}
\left\{
\begin{aligned}
&\hat{C}_k(\mathbf{x}_i) = \left[\frac{1}{k-1}{\sum\limits_{j=1}^{k-1}{\log}\frac{T_k(\mathbf{x}_i)}{T_j(\mathbf{x}_i)}} \right]^{-1}  \\
&\hat{C}_m = \frac{1}{N_i(k_2 - k_ 1+1)}\sum\limits_{i=1}^{N_i}{\sum\limits_{k=k_1}^{k_2}\hat{C}_k(\mathbf{x}_i)}
\end{aligned}
\right.
\end{eqnarray}
where $T_k(\mathbf{x}_i)$ is the distance from $\mathbf{x}_i$ to its $k$-th nearest neighbor.
Interested readers are referred to \cite{levina2004maximum} for more details.

Table \ref{tab:table-estimate-comparison-1} gives the intrinsic dimensions estimated by PCA, MLE, Corr.Dim and NCSC/NCSC-II.
Note the tabulated estimates are given as the average over all classes.
The PCA estimate is an integer for each class but its class average can be fractional while MLE and Corr.Dim only give
fractional estimates for each class.

Since NCSC only gives integer estimates, for comparison purpose, we check the rounded estimates of PCA,
MLE and Corr.Dim, namely $6$ (PCA), $4$ (MLE) and $6$ (Corr.Dim) for the PICS dataset.
For the ORL dataset, the rounded estimates are
$9$ (PCA), $4$ (MLE) and $5$ (Corr.Dim).

It has been widely accepted that the PCA estimator is not satisfactory for nonlinear manifolds, such as
face image datasets. From Table \ref{tab:table-estimate-comparison-1}, it is easy to see that
the estimates of MLE and Corr.Dim are much closer to ours than those of PCA. For the PICS dataset, the
rounded estimate of MLE ($D_\text{c} = 4$) is equal to ours (by NCSC).
For the ORL dataset, the estimate of Corr.Dim ($D_\text{c} = 5$) is closer to
ours ($D_\text{c} = 6$ by NCSC-II) than the estimate of MLE ($D_\text{c} = 4$).

\begin{table}[htb]
\small
\centering
\caption{Comparison of intrinsic dimension estimates by different estimators.}
\begin{threeparttable}[t]
\begin{tabular}{ccccc}
\toprule
\multirow{2}{*}{~~~Dataset~~~}& \multicolumn{4}{c}{~Estimator~} \tabularnewline
\cline{2-5}
&~~~~PCA~~~&~~~~MLE~~~&~~~~Corr.Dim~~~~&  ~~NCSC (proposed)\tabularnewline
\midrule
PICS& $6$&   $4.32$&     $5.55$&  $4$\tnote{~\textdagger}\tabularnewline
%\hline
ORL& $8.9250$& $ 4.34$&  $5.09$&  $6$\tnote{~\textdaggerdbl}\tabularnewline
\bottomrule
\end{tabular}
\label{tab:table-estimate-comparison-1}
\begin{tablenotes}[flushleft]
\item [\textdagger] Estimated by NCSC.
\item [\textdaggerdbl] Estimated by NCSC-II.
\end{tablenotes}
\end{threeparttable}
\end{table}

%\vspace{-1em}
\subsection{Evaluation of NCSC as Classifier}
If the intrinsic dimension of a given dataset is accurately estimated, NCSC/NCSC-II using the estimate
can outperform many of its rivals.

\begin{figure}[htb]
  \centering
  \begin{tabular}{c}
  \includegraphics[width=0.8\textwidth] {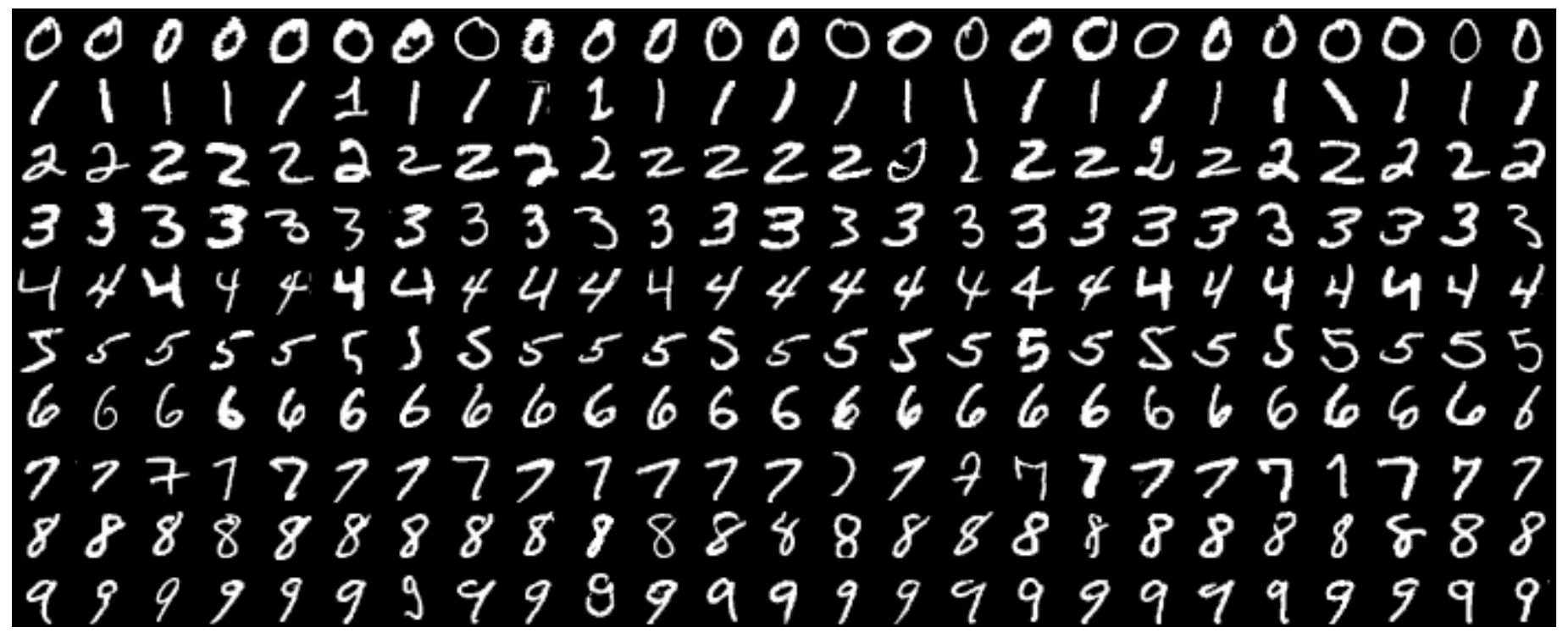}
  \end{tabular}
  \caption{Examples of the MNIST dataset}
  \label{fig:MNIST_examples}
\end{figure}

In this experiment, we compare the classification performance of NCSC/NCSC-II with other related classifiers. The classifiers are evaluated by classifying the samples of the MNIST
dataset. The MNIST dataset of handwritten digits contains $60000$ training images and $10000$ test images \cite{MNIST_database}. The digit images have been size-normalized and centered in a fixed-size image. The image size is $28 \times 28$ pixels.
Figure \ref{fig:MNIST_examples} shows some examples.

Since the MNIST dataset is quite large, we use NCSC-II rather than NCSC to classify the query samples.
In this experiment,
we use the $10\%$ MNIST samples for evaluation.
More specifically, the training set contains the first $6000$ images ($600$ images/class $\times$ $10$ classes) and the query set contains the first $1000$ images ($100$ images/class $\times$ $10$ classes).

In the experiment, a variety of classifiers are used to classify the above mentioned $1000$ query samples.
The accuracies of the different classifiers are compared.
Since the NCSC/NCSC-II framework depends on the intrinsic dimension $D_\text{c}$, we use the estimator of MLE or Corr.Dim to first estimate it.

Table \ref{tab:table-estimate-comparison-2} gives the dimensions estimated from the $10$ training class (each
contains $600$ samples)
by MLE and Corr.Dim. Since currently only the homogenous NCSC classifier is our interest, we have the rounded average estimate of $4$ (over the $10$ classes) by Corr. Dim and $8$ (over the $10$ classes) by MLE.

\begin{table}[htb]
\small
\centering
\caption{Intrinsic dimension estimates by Corr.Dim and MLE.}
\begin{threeparttable}[t]
\begin{tabular}{lllllllllll}
\toprule
Class Index& 1& 2& 3 & 4 & 5 & 6 & 7 & 8 & 9 & 10 \tabularnewline
\midrule
%\hline
%\hline
Corr.Dim{*}& 3.16& 2.58& 4.24& 3.78& 4.08& 4.43& 3.47& 3.54& 3.94& 3.43\tabularnewline
%\hline
MLE\tnote{\textdagger}&6.86~& 4.46~&	9.12~& 8.75~&	8.42~&	7.89~&	7.39~&	7.08~&	8.99~&	7.35 \tabularnewline
\bottomrule
\end{tabular}
\label{tab:table-estimate-comparison-2}
\begin{tablenotes}[flushleft]
\item [*] The rounded average estimate of Corr.Dim over the $10$ classes is $4$.
\item [\textdagger] The rounded average estimate of MLE over the $10$ classes is $8$.
\end{tablenotes}
\end{threeparttable}
\end{table}

To obtain the classification accuracies of related classifiers, we evaluate them on the uncorrupted query data and corrupted query data with the noise
level $\rho$ respectively equal to
$10\%, 20\%$ and $30\%$.

\begin{figure}[htb]
  \centering
  \begin{tabular}{c}
  \includegraphics[width=0.8\textwidth] {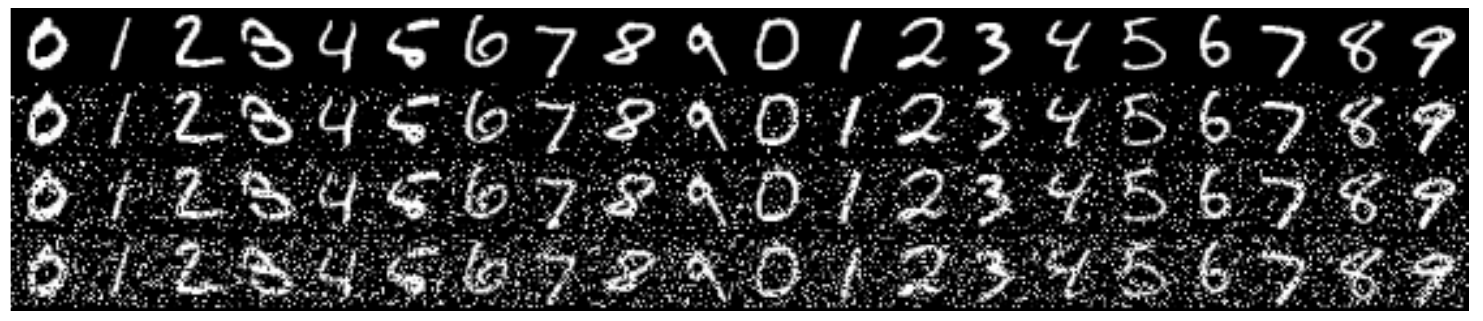}
  \end{tabular}
  \caption{Some MNIST test samples and their corrupted versions with different noise levels. First row: uncorrupted samples.
  Second row: corrupted samples with noise level $\rho = 10\%$.
  Third row: corrupted samples with noise level $\rho = 20\%$.
  Bottom row: corrupted samples with noise level $\rho = 30\%$.}
  \label{fig:MNIST_examples_currupted}
\end{figure}

The corrupted pixels are uniformly chosen in a target image. Since an MNIST image sample contains $28 \times 28 = 784$ pixels, for the noise level
$\rho$,
the number of corrupted pixels is a rounded integer of $784 \times \rho$. The intensities of the corrupted pixels are
uniformly distributed in $\{i_\text{min}, \cdots,i_\text{max}\}$, where $i_\text{min}$ and $i_\text{max}$ are respectively
the minimum and maximum intensity of the uncorrupted image.
Figure \ref{fig:MNIST_examples_currupted} gives some MNIST samples and their corrupted versions.

Table \ref{tab:classification-accuracy-comparison} gives the accuracies  of NN, NFL, NS and NCSC-II with $D_\text{c}$ respectively equal to $1$, $4$ and $8$
for classifying the $1000$ query samples ($100$ query samples/class $\times$ $10$ classes) respectively corrupted by different noise levels.

Note each training class contains $600$ samples, which in fact are too many for NFL to deal with.
In order to obtain the distance of a query sample to a given class,
NFL computes the projection distances of a query sample to ${\binom {600} 2} = 179700$ feature lines.
The computational cost is very high.
Thus, we could not obtain in an acceptable time the classification accuracy of NFL, which happens to be NCSC with $D_\text{c} = 1$.
For comparison purpose, we give the classification accuracy of NCSC-II with $D_\text{c} = 1$, which is the fast version of NFL.

We are particularly interested in the classification accuracies of NCSC-II using the $D_\text{c}$ respectively estimated by Corr.Dim and MLE, i.e., $D_\text{c} = 4$ and $D_\text{c} = 8$.
Table \ref{tab:classification-accuracy-comparison} shows the results using the estimates by Corr.Dim and MLE.
It is easy to see that
NCSC-II using the estimate by MLE yields better
classification accuracies than those associated with Corr.Dim.
It also shows that the highest accuracy ($96.0\%$) is obtained by NCSC-II with the estimate by MLE ($D_\text{c} = 8$) on the  uncorrupted MNIST data samples.

From this point of view, we contend that \emph{for the MNIST dataset, the estimate by MLE is better than the estimate
by Corr.Dim.}

The NS algorithm fails in this experiment, yielding the classification accuracies not larger than $54.7\%$. As mentioned in previous sections, NS can be viewed as
a high dimensional extreme of NCSC and NCSC-II.
In this experiment on the MNIST dataset ($600 \times 10$ training samples), the $10$ subspaces (corresponding to the $10$ training classes)
employed by NS are intrinsically \textcolor{black}{nearly} $600$ dimensional\footnote{\textcolor{black}{The intrinsic dimension is equal to the rank of the matrix whose columns are the 600 training vectors of a target class.}}.
The intrinsic dimension is unnecessarily high
and there is a high probability that the various subspaces have significant overlaps.
See Expression (\ref{eq:nest}) for more on this issue.

Furthermore, note, as another extreme, NN yields the second lowest accuracy in this experiment.

\begin{table}[htbp]
  \small
  \centering
  \caption{Comparison of classification accuracies of several subspace-based classifiers.}\vspace{0.5em}
  \begin{threeparttable}[t]
    \begin{tabular}{c|l|llll}
    \toprule
    \multicolumn{2}{c|}{Noise Level} & ~~~0\%~~~~~~~~     & 10\%~~~~~~~~    & 20\%~~~~~~    & 30\%~~~    \\
    %\hline
    %\hline
    \midrule
    \multicolumn{2}{c|}{NN} & ~~~92.3\%    & 91.6\%    & 91.9\%    & 91.6\%\\
    \hline
    \multicolumn{2}{c|}{NFL~\tnote{\#}} & \multicolumn{4}{c}{\textbf{---}} \\
    \hline
    \multirow{2}[7]{*}{NCSC-II} & $D_\text{c} = 1$~\tnote{\textdagger}  & ~~~93.4\%    & 93.0\%    & 93.4\%    & 92.1\%\\
\cline{2-6}          & $D_\text{c} = 4$ (Corr.Dim) & ~~~94.9\%    & 94.1\%    & 94.0\%    & 91.9\%\\
\cline{2-6}          & $D_\text{c} = 8$ (MLE)~{*}& ~~~96.0\%~\tnote{\textdaggerdbl}    & 94.6\%    & 94.0\%    & 92.1\%\\
    \hline
    \multicolumn{2}{c|}{NS} & ~~~54.7\%    & 12.7\%    & 12.5\%    & 10.5\%  \\
    %\hline
    \bottomrule
    \end{tabular}%\label{tab:classification-accuracy-comparison}
  \begin{tablenotes}[flushleft]
    \item [\#] Unable to obtain experimental results in an acceptable time.
    \item [\textdagger] Corresponding to the fast NFL classifier, i.e., NCSC-II with $\kappa = 2$.
    \item [*] Corresponding to the optimal classification performance, using $D_\text{c}$ estimated by MLE.
    \item [\textdaggerdbl] The highest classification accuracy in this experiment.
  \end{tablenotes}
  \end{threeparttable}
\label{tab:classification-accuracy-comparison}
\end{table}

Table \ref{tab:classification-accuracy-comparison} supports to our claim that with appropriately estimated
dimension parameter, NCSC (or NCSC-II as demonstrated in this experiment) as a parameterized classifier, can outperform its rivals.
Its classification accuracy depends on the accuracy of the estimated dimension.

%\vspace{-3em}
\subsection{Comparison of Run Time}
Table \ref{tab:run-time} gives the comparison of the run time of classifying one query sample by a variety of classifiers, including NN, NFL NS and NCSC-II, respectively on the MNIST data with a small training
set of $6000$ images, (i.e., $600~\text{images/class}~\times~10~\text{classes}$) and a large training set of $60000$ images \textcolor{black}{(i.e., the complete MNIST
training set of $10$ classes, each approximately having $6000$ images)}.\footnote{
The environment for this experiment is MATLAB on an x64 PC with 32GB memory and an Intel
CPU at 3.2 GHz.}

\begin{table}[htbp]
%\begin{table}[H]
  \small
  \centering
  \caption{Comparison of run time of different classifiers.}\vspace{0.5em}
  \begin{threeparttable}[t]
    \begin{tabular}{ccccccc}
    \toprule
    \multirow{3}[0]{*}{Traing Set Size} & \multicolumn{6}{c}{Run Time (second)} \\
    \cline{2-7}
          & \multirow{2}[0]{*}{NN} & \multirow{2}[0]{*}{NS} & \multirow{2}[0]{*}{NFL} & \multicolumn{3}{c}{NCSC-II} \\
    \cline{5-7}
          &       &       &       & $D_\text{c} = 1$ & $D_\text{c} = 4$ & $D_\text{c} = 8$ \\
    \midrule
    $6000$   & 0.05 & 1.53 & $~~~> 35~\text{minutes}$ & 7.03 & 8.47 & 8.64 \\
    $60000$ & 0.49 & 10.70 & $>50~\text{hours}$ & 70.70 & 83.51 & 88.32 \\
    \bottomrule
    \end{tabular}%
    \end{threeparttable}
\label{tab:run-time}
\end{table}%

The classifications on the large training set yield a similar classification accuracy comparison
to that given in Table \ref{tab:classification-accuracy-comparison}, where the classification accuracies of NN and NS are less than those of NCSC respectively
with $D_\text{c} = 1, 4$ and $8$.\footnote{
NS fails to accurately classify query samples, yielding a classification accuracy of approximate $55\%$ for the small training set and
approximate $23\%$ for the large training set. }

The run time in Table \ref{tab:run-time} is given as the average time of $100$ independent classifications.\footnote{In the run time experiments for NCSC-II, we assume that the neighborhood information of each training sample is obtained before classification.}

Among the evaluated classifiers,
NFL  is the most time-consuming. The time of classifying a query sample by NFL is larger than $35$ minutes with the small training
set and $50$ hours with the large training set.

Using the small training set, NCSC (with $D_\text{c} = 1,4$ and $8$) yields moderate run time between $7-9$ seconds.
Using the large training set, whose size is $10$ times that of the small training set,
NCSC-II yields the run time between $70-90$ seconds.\footnote{
With the help of parallel computing, the run time can be significantly reduced
for practical uses. }
It is shown that the time complexity of NCSC-II is proportional to the size of training set.

\section{Conclusion}\label{section:conclusion}

We propose a novel classifier, called NCSC, and a fast version NCSC-II. The proposed algorithms can be used as an intrinsic dimension estimator for a
labeled dataset or a classifier when the dimension parameter is given.

The proposed NCSC/NCSC-II generalizes some classical classifiers including NN (Nearest Neighbor), NFL (Nearest Feature Line)
and has a close relation to NS (Nearest Subspace). We proved that NN and NFL are two special cases of low-dimensional
NCSC (with the dimension parameter $D_\text{c}$ respectively equal to $0$ and $1$) while NS is closely associated with the highest dimensional NCSC.

The NCSC framework is a constrained linear regression problem under the least squares principle.
First, the regression coefficients are regularized to ensure that the sum of all coefficients is equal to $1$.
Second, the regression coefficient vector is sparse under the assumption that the data manifold has a relatively low dimension.

We use $\ell_0$-norm to ensure the sparsity of the coefficient vector. Then, we define the constrained subspace dimension given the bound on the $\ell_0$-norm of coefficient vector. We contend that as an approximation to the target data manifold,
the constrained subspaces should all have a dimension equal to the manifold dimension in order to approximate the manifold more accurately.

In order to reduce the computational complexity of NCSC, which depends heavily on the number of subsets of the training samples, we further propose a fast version of NCSC, called NCSC-II. Under the assumption that the nearest neighbors of a target data point of the same class can capture the local intrinsic dimension,
NCSC-II employs a $\kappa$-neighbors representation to model the sparsity of coefficient vector. Using the $\kappa$-neighbors representation, the computational complexity is significantly reduced.

We evaluate NCSC/NCSC-II respectively as an estimator and a classifier on several publicly available
datasets and compare the results of NCSC/NCSC-II with those obtained by the MLE, Corr.Dim estimators and the NN, NFL, NS classifiers.

Experiments show NCSC/NCSC-II can serve either as an estimator or a classifier with good performance. NCSC/NCSC-II can serve as a benchmark in terms of classification accuracy to evaluate the performances of other estimators . On the other hand, when an appropriate dimension parameter is given, NCSC/NCSC-II outperforms a wide range of its subspace-based rivals.

\section*{Acknowledgments}
This work was partly supported by the National Natural Science Foundation of China.


\begin{thebibliography}{10}
\providecommand{\url}[1]{#1}
\csname url@samestyle\endcsname
\providecommand{\newblock}{\relax}
\providecommand{\bibinfo}[2]{#2}
\providecommand{\BIBentrySTDinterwordspacing}{\spaceskip=0pt\relax}
\providecommand{\BIBentryALTinterwordstretchfactor}{4}
\providecommand{\BIBentryALTinterwordspacing}{\spaceskip=\fontdimen2\font plus
\BIBentryALTinterwordstretchfactor\fontdimen3\font minus
  \fontdimen4\font\relax}
\providecommand{\BIBforeignlanguage}[2]{{%
\expandafter\ifx\csname l@#1\endcsname\relax
\typeout{** WARNING: IEEEtran.bst: No hyphenation pattern has been}%
\typeout{** loaded for the language `#1'. Using the pattern for}%
\typeout{** the default language instead.}%
\else
\language=\csname l@#1\endcsname
\fi
#2}}
\providecommand{\BIBdecl}{\relax}
\BIBdecl

\bibitem{manifolds_handwritten_digits}
G.~E. Hinton, P.~Dayan, and M.~Revow, ``Modeling the manifolds of images of
  handwritten digits,'' \emph{IEEE Transactions on Neural Networks}, vol.~8,
  pp. 65--74, 1997.

\bibitem{Eigenfaces_recognition}
M.~Turk and A.~Pentland, ``Eigenfaces for recognition,'' \emph{Journal of
  Cognitive Neuroscience}, vol.~3, no.~1, pp. 71--86, 1991.

\bibitem{Whitney_Reduction_Network}
D.~Broomhead and M.~Kirby, ``The {W}hitney reduction network: A method for
  computing autoassociative graphs,'' \emph{Neural Computation}, vol.~13,
  no.~11, pp. 2595--2616, 2001.

\bibitem{isomap2000}
J.~Tenenbaum, V.~Silva, and J.~Langford, ``A global geometric framework for
  nonlinear dimensionality reduction,'' \emph{Science}, vol. 290, no. 5500, pp.
  2319--2323, 2000.

\bibitem{donoho2003hessian}
D.~L. Donoho and C.~Grimes, ``Hessian eigenmaps: Locally linear embedding
  techniques for high-dimensional data,'' \emph{Proceedings of the National
  Academy of Sciences of the United States of America}, vol. 100, no.~10, p.
  5591, 2003.

\bibitem{LLE2000}
S.~Roweis and L.~Saul, ``Nonlinear dimensionality reduction by locally linear
  embedding,'' \emph{Science}, vol. 290, no. 5500, pp. 2323--2326, 2000.

\bibitem{weinberger2004unsupervised}
K.~Weinberger and L.~Saul, ``Unsupervised learning of image manifolds by
  semidefinite programming,'' in \emph{Proceedings of the 2004 IEEE Computer
  Society Conference on Computer Vision and Pattern Recognition, 2004. CVPR
  2004.}, vol.~2.\hskip 1em plus 0.5em minus 0.4em\relax IEEE, 2004, pp.
  II--988.

\bibitem{zhang2004principal}
Z.~Zhang and H.~Zha, ``Principal manifolds and nonlinear dimensionality
  reduction via tangent space alignment,'' \emph{Journal of Shanghai University
  (English Edition)}, vol.~8, no.~4, pp. 406--424, 2004.

\bibitem{Belkin01laplacianeigenmaps}
M.~Belkin and P.~Niyogi, ``Laplacian eigenmaps and spectral techniques for
  embedding and clustering,'' in \emph{Advances in Neural Information
  Processing Systems 14}.\hskip 1em plus 0.5em minus 0.4em\relax MIT Press,
  2001, pp. 585--591.

\bibitem{Li99facerecognition}
S.~Z. Li and J.~Lu, ``Face recognition using the nearest feature line method,''
  \emph{IEEE Transactions on Neural Networks}, vol.~10, pp. 439--443, 1999.

\bibitem{Solve_constrained001}
T.~Coleman and Y.~Li, ``A reflective newton method for minimizing a quadratic
  function subject to bounds on some of the variables,'' \emph{SIAM Journal on
  Optimization}, vol.~6, no.~4, pp. 1040--1058, 1996.

\bibitem{Solve_constrained002}
W.~M. Gill~P.E. and M.~Wright, \emph{Practical Optimization}.\hskip 1em plus
  0.5em minus 0.4em\relax London, UK: Academic Press, 1981.

\bibitem{levina2004maximum}
E.~Levina and P.~J. Bickel, ``Maximum likelihood estimation of intrinsic
  dimension,'' in \emph{Advances in Neural Information Processing Systems 17},
  L.~K. Saul, Y.~Weiss, and l.~Bottou, Eds., Cambridge, MA, December 2005, pp.
  777--784.

\bibitem{carter2010local}
K.~Carter, R.~Raich, and A.~Hero, ``On local intrinsic dimension estimation and
  its applications,'' \emph{IEEE Transactions on Signal Processing}, vol.~58,
  no.~2, pp. 650--663, 2010.

\bibitem{PICS_database}
Stirling, ``Psychological image collection at {S}tirling ({PICS}),''
  http://pics.psych.stir.ac.uk/2D\_face\_sets.htm.

\bibitem{orl_database}
AT\&T, ``The database of faces,'' 2002~~~~~~~~,
  http://www.cl.cam.ac.uk/research/dtg/attarchive/facedatabase.html.

\bibitem{grassberger1983measuring}
P.~Grassberger and I.~Procaccia, ``Measuring the strangeness of strange
  attractors,'' \emph{Physica D: Nonlinear Phenomena}, vol.~9, no.~1, pp.
  189--208, 1983.

\bibitem{camastra2002estimating}
F.~Camastra and A.~Vinciarelli, ``Estimating the intrinsic dimension of data
  with a fractal-based method,'' \emph{IEEE Transactions on Pattern Analysis
  and Machine Intelligence}, vol.~24, no.~10, pp. 1404--1407, 2002.

\bibitem{fukunaga1971algorithm}
K.~Fukunaga and D.~Olsen, ``An algorithm for finding intrinsic dimensionality
  of data,'' \emph{IEEE Transactions on Computers}, vol. 100, no.~2, pp.
  176--183, 1971.

\bibitem{bruske1998intrinsic}
J.~Bruske and G.~Sommer, ``Intrinsic dimensionality estimation with optimally
  topology preserving maps,'' \emph{IEEE Transactions on Pattern Analysis and
  Machine Intelligence}, vol.~20, no.~5, pp. 572--575, 1998.

\bibitem{massoud2007manifold}
A.~massoud Farahmand, C.~Szepesv{\'a}ri, and J.~Audibert, ``Manifold-adaptive
  dimension estimation,'' in \emph{Proceedings of the 24th international
  conference on Machine learning}, 2007, pp. 265--272.

\bibitem{fan2009intrinsic}
M.~Fan, H.~Qiao, and B.~Zhang, ``Intrinsic dimension estimation of manifolds by
  incising balls,'' \emph{Pattern Recognition}, vol.~42, no.~5, pp. 780--787,
  2009.

\bibitem{MNIST_database}
Y.~LeCun and C.~Cortes, ``The {MNIST} database of handwritten digits,''~\\
  http://yann.lecun.com/exdb/mnist/.

\end{thebibliography}
\end{document}